\documentclass[table]{article} % For LaTeX2e
\usepackage{iclr2025_conference,times}

\usepackage{amsmath,amsfonts,bm}

\def\eqref#1{equation~\ref{#1}}
\def\1{\bm{1}}

\DeclareMathAlphabet{\mathsfit}{\encodingdefault}{\sfdefault}{m}{sl}
\SetMathAlphabet{\mathsfit}{bold}{\encodingdefault}{\sfdefault}{bx}{n}

\usepackage[utf8]{inputenc} % allow utf-8 input
\usepackage[T1]{fontenc}    % use 8-bit T1 fonts
\usepackage{hyperref}       % hyperlinks
\usepackage{url}  
\usepackage{booktabs}       % professional-quality tables
\usepackage{amsfonts}       % blackboard math symbols
\usepackage{nicefrac}       % compact symbols for 1/2, etc.
\usepackage{microtype}      % microtypography
\usepackage{amsmath}
\usepackage{graphicx}
\usepackage{wrapfig}
\usepackage{subcaption}
\usepackage{bm}
\usepackage{array}
\usepackage{mathtools}
\usepackage{relsize} 

\newcommand{\ours}{EqNIO}

\title{\ours{}: Subequivariant Neural Inertial \\Odometry}

\author{Royina Karegoudra Jayanth\thanks{denotes the equal contribution},\,\, Yinshuang Xu\footnotemark[1],\,\, Ziyun Wang, \,\,Evangelos Chatzipantazis, \\\textbf{Daniel Gehrig, \,\, Kostas Daniilidis}\\\\
University of Pennsylvania\\\\
\texttt{\{royinakj,xuyin,ziyunw,vaghat,dgehrig\}@seas.upenn.edu}\\ \texttt{kostas@cis.upenn.edu}
}

\iclrfinalcopy % Uncomment for camera-ready version, but NOT for submission.
\begin{document}

\maketitle
\vspace{-5mm}
\begin{abstract}
Neural network-based odometry using accelerometer and gyroscope readings from a single IMU can achieve robust, and low-drift localization capabilities, through the use of \emph{neural displacement priors}. These priors learn to produce denoised displacement measurements but need to ignore data variations due to specific IMU mount orientation and motion directions, hindering generalization.
This work introduces EqNIO, which addresses this challenge with \emph{canonical displacement priors}. We train an off-the-shelf architecture with IMU measurements that are mapped into a canonical gravity-aligned frame with learnable yaw. The outputs (displacement and covariance) are mapped back to the original frame.
% We first map IMU measurements into a canonical gravity-aligned frame with a learnable yaw, then run an off-the-shelf network, and map the outputs (displacement and covariance) back into the original frame. 
To maximize generalization, we find that these learnable yaw frames must transform equivariantly with global trajectory rotations and reflections across the gravity direction, \emph{i.e.} action by the roto-reflection group $O_g(3)$ which preserves gravity (a subgroup of $O(3)$). This renders the displacement prior $O(3)$ \emph{subequivariant}.
We tailor specific linear, convolutional and non-linear layers that commute with the actions of the group. 
Moreover, we introduce a bijective decomposition of angular rates into vectors that transform similarly to accelerations, allowing us to leverage both measurements types. Natively, angular rates would need to be inverted upon reflection, unlike acceleration, which hinders their joint processing.
We highlight EqNIO's flexibility and generalization capabilities by applying it to both filter-based (TLIO), and end-to-end (RONIN) architectures, and outperforming existing methods that use \emph{soft} equivariance from auxiliary losses or data augmentation on the TLIO, Aria, RONIN, RIDI and OxIOD datasets. We believe this work paves the way to low-drift, and generalizable neural inertial odometry on edge-devices.

\end{abstract}
\begin{figure}[ht]
\vspace{-5mm}
\begin{center}
\centerline{\includegraphics[width=\linewidth]{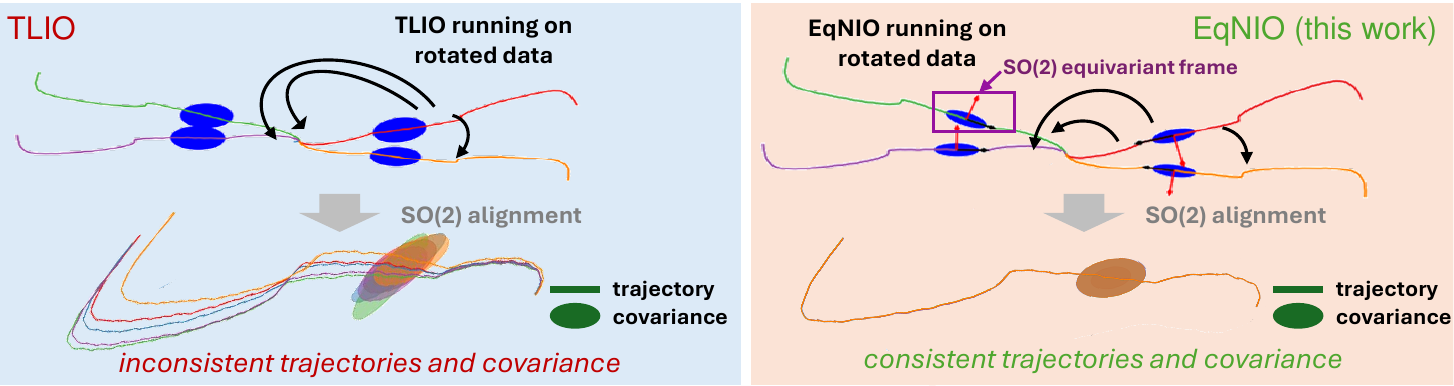}}
%\vspace{-3mm}
\caption{Predicted trajectories and covariance ellipsoids from TLIO (left) and subequivariant \ours{} (right) for 5 identical trajectories with different IMU frames. The de-rotated trajectories and ellipsoids of TLIO demonstrate significant inconsistency, while the ones by \ours{} are perfectly aligned.}
%Our \ours{} predicts equivariant frames that are aligned with the covariance ellipsoids.
%The frames predicted by \ours{} are equivariant and aligned to the principal axis of covariance ellipsoid.}
\label{fig:teaser}
\end{center}
\vspace{-10mm}
\end{figure}

\section{Introduction}
\vspace{-2mm}
Inertial Measurement Units (IMUs) are commodity sensors that measure the forces and angular velocities experienced by a body. Due to their low cost, they are widely used in diverse applications such as robot navigation and Mixed Reality where they facilitate the precise and rapid tracking of body frames. However, since they are differential sensors, relying only on IMUs invariably results in drift. 
Traditional Visual Inertial Odometry (VIO) approaches can effectively mitigate this drift by combining IMU measurements with features extracted from camera images. Still, these images are of limited use in high-speed scenarios with challenging lighting conditions, since they can suffer from saturation, and blurring artifacts. Recently, a novel class of methods has emerged that instead mitigates this drift with statistical motion priors that are directly learned from IMU data alone~\citep{liu2020tlio, herath2020ronin}. These methods perform competitively with VIO methods despite using only a single IMU sensor. However, learning generalizable priors proves challenging: while identifying specific motion patterns, they must learn to ignore data differences due to particular IMU mount orientations, and the direction of the motion patterns. However, they often fail to do so in practice (see Fig.~\ref{fig:teaser}, left state-of-the-art method TLIO~\citep{liu2020tlio}), yielding large trajectory variations when observing simply rotated input data. They do this despite using data augmentation strategies that include random data rotations.
In this work, we introduce \textbf{EqNIO}, which simplifies this task by learning \emph{canonical displacement priors}. These priors are invariant to IMU orientation and are easier to learn, more robust, and, as a result, more generalizable than previous work. EqNIO achieves this by first mapping sequences of IMU measurements into a canonical, gravity-aligned frame $F$, then passing them to a neural network, and finally mapping the outputs back to the original frame. 
Our design can flexibly integrate arbitrary off-the-shelf methods such as TLIO~\citep{liu2020tlio} and RONIN~\citep{herath2020ronin}, provided that suitable change of basis maps are defined for the network outputs (displacement and covariance of TLIO, and only linear velocity for RONIN). 
We show that IMU data can be effectively canonicalized, i.e., projected into an invariant space, in two steps: We first perform gravity-alignment as done traditionally in inertial odometry, \emph{i.e.} apply simple rotations that align the $z$-axis with a known gravity direction, and then perform a yaw roto-reflection into a canonical yaw orientation. While the gravity direction can be easily estimated from an off-the-shelf Extended Kalman Filter (EKF)~\citep{liu2020tlio}, canonical yaw orientations cannot be observed, and must thus be learned from data. Moreover, this frame must generalize across data collected under arbitrary rotations around gravity, and reflections across planes parallel to gravity.  

We specifically design a model that predicts an equivariant frame (Eq. Frame) that roto-reflects equivariantly with the input and achieves inherent generalization via \emph{equivariant processing}. Our equivariant model, \textbf{EqNIO}, takes in sequences of IMU gyroscope and accelerometer measurements and processes them in a way that commutes with the action of roto-reflections around gravity. To achieve this we introduce suitable unique preprocessing steps which map both accelerometer and gyroscope measurements into a space that transforms consistently under the group action. We identify these roto-reflections are elements of the group $O_{g}(3)$, a subgroup of the orthogonal group $O(3)$ which preserves the gravity direction, and is isomorphic to $O(2)$. This isomorphism allows us to specifically tailor $O(2)$ equivariant MLPs, convolutions, and non-linear layers, which, due to the subgroup property are also called $O(3)$ \emph{subequivariant}. 

\textbf{Contributions:} In summary, our contributions are: \emph{(i)} We introduce a canonicalization scheme that leverages gravity and an estimated subequivariant frame to map IMU measurements into a canonical orientation. This procedure can be flexibly applied to arbitrary off-the-shelf network architectures by mapping the inputs into the canonical space, and mapping the outputs back into the original space. Moreover, it simplifies the learning of displacement priors and produces inherently more robust, and generalizable results than previous work. \emph{(ii)} We formalize the symmetry properties of IMU measurements under arbitrary roto-reflections from $O_{g}(3)$, and derive unique preprocessing steps which map both accelerometer and gyroscope measurements into a space in which these roto-reflections act consistently. We further split these vectors into 2D $O(2)$ equivariant vector features, and 1D \emph{invariant} scalar. \emph{(iii)} To process this data, we specifically tailor an $O(2)$ equivariant network to regress canonical yaw frames. It leverages specialized $O(2)$ equivariant MLPs and convolution to process vector features, conventional layers to process scalar features, and equivariant non-linearities that facilitate the interaction of vector and scalar features. 

We demonstrate the generality of our framework by applying it to two neural inertial odometry methods, TLIO~\citep{liu2020tlio}, and RONIN~\citep{herath2020ronin}, and performing detailed quantitative and qualitative analysis comparing EqNIO with several state-of-the-art methods across diverse datasets. This evaluation confirms the superiority of our approach. Our method establishes a new state-of-the-art performance in inertial-only odometry, significantly enhancing the accuracy, reliability, and generalization of existing methods.

\vspace{-4.5mm}
\section{Related Work}
\vspace{-3mm}
\textbf{Model-based Inertial Odometry:} Purely Inertial Odometry can be broadly classified into two categories: kinematics-based and learning-based approaches. Kinematics-based approaches~\citep{Leishman_StateEstimation,titterton2004strapdown, Bortz1971} leverage analytical solutions based on double integration that suffer from drift accumulation over time when applied to consumer-grade IMUs. To mitigate this drift, loop closures~\citep{solin2018inertial}, and other pseudo measurements derived from IMU data that are drift-free have been explored ~\citep{groves2015principles,hartley2020contact,brajdic2013walk}. In the context of  Pedestrian Dead Reckoning~\citep{jimenez2009comparison}, these include step counting~\citep{ho2016step,brajdic2013walk}, detection of the system being static~\citep{foxlin2005pedestrian,rajagopal2008personal} and gait estimation~\citep{beaufils2019robust}. 

\textbf{Learning-based Inertial Odometry:} Recently, RIDI~\citep{Yan_2018_ECCV}, PDRNet~\citep{pdrnet} and RONIN~\citep{herath2020ronin} proposed learning-based approaches using CNNs, RNNs, and TCNs that regress velocity. RIDI uses the regressed velocity to correct the IMU measurements, while RONIN 
directly integrates the regressed velocities but assumes orientation information. Denoising networks either regress the IMU biases~\citep{9119813,deepimubias,aiimu} or output the denoised IMU measurements~\citep{9811989}. While \citet{deepimubias} uses constant covariance, AI-IMU~\citep{aiimu} estimates the covariance for automotive applications. Displacement-based neural network methods like IONet~\citep{chen2018ionet}, TLIO~\citep{liu2020tlio}, RNIN-VIO~\citep{chen2021rnin}, and IDOL~\citep{sun2021idol} directly estimate 2D/3D displacement. Unlike TLIO, where the neural network also regresses a diagonal covariance matrix,~\citet{russell2021multivariate} parameterize the full covariance matrix using Pearson correlation. RNIN-VIO extends TLIO's method to continuous human motion adding a loss function for long-term accuracy. Unlike these methods, EqNIO learns canonical displacement priors and thus generalizes better to arbitrary IMU orientation and motion directions.

\textbf{Equivariant Inertial Odometry:} The previous learning-based approaches~\citep{liu2020tlio,chen2021rnin} use $SO(2)$ augmentation strategies to achieve approximate $SO(2)$ equivariance.  MotionTransformer~\citep{chen2019motiontransformer} used GAN-based RNN encoder to transfer IMU data into domain-invariant space by separating the domain-related constant.
Recently, RIO~\citep{cao2022rio} demonstrated the benefits of approximate $SO(2)$ equivariance with an auxiliary loss, introduced Adaptive Test Time Training (TTT), and uncertainty estimation via ensemble of models. We propose integrating strict equivariance by design directly into the framework. Additionally, no prior work has addressed reflection equivariance, which requires specific preprocessing of gyroscope data for it to adhere to the right-hand rule. Our novel $O(2)$ equivariant framework can be seamlessly integrated with existing learning-based inertial navigation systems, showing benefits on TLIO and RONIN.

\textbf{Equivariant Networks: } Group equivariant networks~\citep{cohen2016group} provide deep learning pipelines that are 
equivariant by design with respect to group transformations of the input. Extensive research has been conducted on how these networks process a variety of inputs, including point clouds~\citep{thomas2018tensor,chen2021equivariant,deng2021vector}, 
2D~\citep{worrall2017harmonic,weiler2019general}, 
3D~\citep{weiler20183d, esteves2019equivariant}, 
and spherical images~\citep{cohen2018spherical,esteves2018learning,esteves2020spin,esteves2023scaling}, 
graphs~\citep{satorras2021n} and general manifolds~\citep{cohen2019general, cohen2019gauge, weiler2021coordinate, xu2024se}.
However, there is no current applicable equivariant model for the prediction of displacement prior using IMU data which is a sequence of vectors and scalars. We are the first to apply equivariant models for neural integration of IMUs. On the other hand, general theories and methods have been developed for networks that are equivariant to $E(n)$ and its subgroups.~\citet{cesa2021program,xu2022unified} utilize Fourier analysis to design steerable kernels in CNNs on homogeneous space, while~\citet{finzi2021practical} proposed a numerical algorithm to compute a kernel by solving the linear equivariant map constraint.~\cite{villar2021scalars} demonstrated that any $O(n)$ equivariant function can be represented using a set of scalars and vectors. However, applying these to neural integration of IMUs is not straightforward as gravity's presence introduces subequivariance, angular velocity in the input data follows the right-hand rule, and the input is a sequence with a time dimension. Related 
approaches~\citep{han2022learning,chen2023subequivariant} tackle subequivariance using equivariant graph networks and calculating gram matrices achieving simple $O(2)$ equivariance. However, dealing with data that obey the right-hand rule (\emph{e.g.} angular rates), has been underexplored, and is addressed in the current work. 

\vspace{-3mm}
\section{Problem Setup}
\vspace{-1mm}
\label{problem_setup}
\begin{figure}[!t]
     \centering
    % First row
    \includegraphics[width=0.95\linewidth]{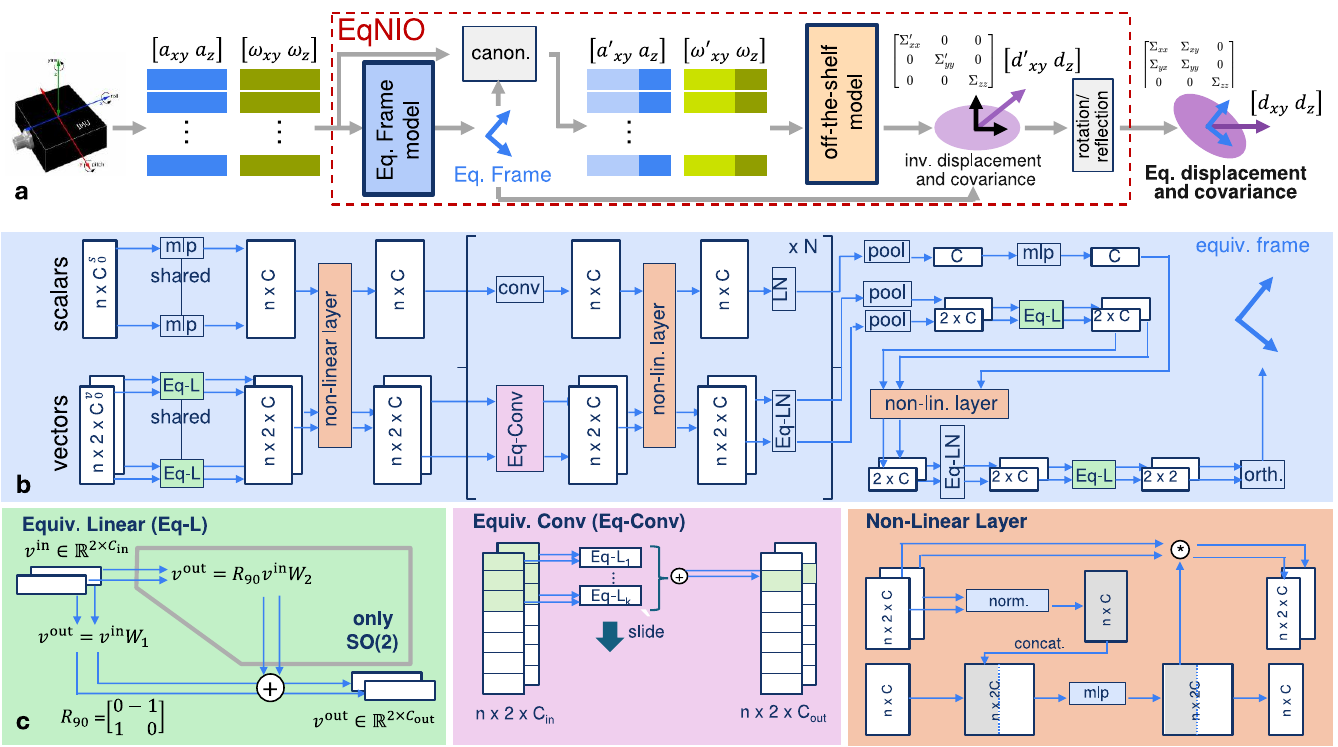}
    \vspace{-3mm}
    \caption{ 
    EqNIO (a) processes gravity-aligned IMU measurements, 
    $\{(a_i,\omega_i)\}_{i=1}^n$. 
    An equivariant network (blue) predicts a canonical equivariant frame $F$ into which IMU measurements are mapped, \emph{i.e. canonicalized}, yielding invariant inputs $\{(a'_i, \omega'_i)\}_{i=1}^n$. 
    A conventional neural network then predicts invariant displacement ($d'$) and covariance ($\Sigma'$) which are mapped back yielding equivariant displacement ($d$) and covariance ($\Sigma$). 
    The equivariant network (b) takes as input $n\times C_0^s$ scalars, and $n\times C_0^v$ vectors:
    Vectors are processed by equivariant layers (Eq-L, Eq-Conv, Eq-LN), while scalars are separately processed with conventional layers. 
    Eq-L (green) uses two weights $W_1,W_2$ for SO(2) equivariance, and only $W_1$ for O(2) equivariance. Eq-Conv (pink) uses Eq-L to perform 1-D convolutions over time. The equivariant non-linear layer (orange) mixes vector and scalar features.}
    \vspace{-0.09cm}
    \label{pipeline}
    \vspace{-4mm}
\end{figure}
This paper targets neural inertial odometry using data from a single IMU, comprised of an accelerometer (giving linear accelerations $a_i\in\mathbb{R}^3$) and gyroscope (giving angular velocity $\omega_i\in\mathbb{R}^3$). IMU's measure sequences of data $\{(a_i, \omega_i)\}_{i=1}^n$, each expressed in the local IMU body frame, $b$, at time $t_i$.
These are related to the true IMU acceleration $\bar{a}_i$ and angular rates $\bar{\omega}_i$ via 
{\small
\begin{align}
    \quad\omega_i = \bar{\omega}_i + b_i^g + \eta^g_i\quad
   \quad a_i = \bar{a}_i - \prescript{w}{b}{R_i}^T  g + b_i^a + \eta^a_i\quad
\end{align}
}
where $g$ is gravity pointing downward in world frame $w$, $\prescript{w}{b}{R_i}$ is the transformation between $b$ and $w$ at time $t_i$, and $b^g_i,b^a_i$ and $\eta^g_i,\eta^a_i$ are IMU biases and noises respectively. Naively integrating angular rates and accelerations to get positions $p_i$ and orientations $\prescript{w}{b}{R_i}$ leads to significant drift due to sensor noise and unknown biases. We thus turn our attention to neural displacement priors $\Phi$, which regress accurate 2D linear velocities~\citep{herath2020ronin} or 3D displacement $d\in\mathbb{R}^3$ and covariances $\Sigma\in\mathbb{R}^{3\times 3}$~\citep{liu2020tlio} from sequences of IMU measurements, and thus have the form 
{\small
\begin{align}
    d, \Sigma &= \Phi\left(\{(a_i, \omega_i)\}_{i=1}^n\right)
\end{align}
}
where $d\in\mathbb{R}^3$ denotes displacement on the time interval $[t_1,t_n]$, and $\Sigma\in\mathbb{R}^{3\times 3}$ denotes associated covariance prediction. For instance, \cite{liu2020tlio}, use these network predictions as measurements and fuses them in an EKF estimating the IMU state in $w$, \emph{i.e.} orientation, position, velocity, and IMU biases. Preliminaries on terms used in inertial odometry are included in App.~\ref{inertial_odomerty_pre} and details of EKF and IMU measurement model are included in App.~\ref{App:EKF}.

% To reduce the data variability in IMU measurements, we map them to a \emph{gravity-aligned frame} (i.e. perform \emph{gravity-alignment}), which has its z-axis aligned with the gravity direction. We do this by simple rotations. Typically, the gravity direction can be estimated from an auxiliary \emph{visual-inertial} odometry (VIO) method or can be estimated from the dominant direction of accelerometer readings at rest. Reducing the data variability aids in learning priors associated with particular IMU motions, and also simplifies odometry due to the elimination of two degrees of freedom. This frame is, however, ill-defined, since simply rotating it around z or reflecting it across planes parallel to z (applications of rotations or roto-reflections from the groups $SO(2)$ and $O(2)$) result in new valid gravity-aligned frames. We train a network $\Phi$ that takes as input a sequence of $n$ IMU measurements ($a_i, \omega_i \in\mathbb{R}^3$) in a gravity-aligned frame and regresses 3D displacement ($d\in\mathbb{R}^3$) and covariance ($\Sigma\in\mathbb{R}^{3\times 3}$) in the same frame, and thus 
% \begin{equation}
% d, \Sigma = \Phi(\{a_i, \omega_i\}_{i=1}^n)
% \end{equation}

\begin{figure*}
    \centering
    \includegraphics[width=1\linewidth]{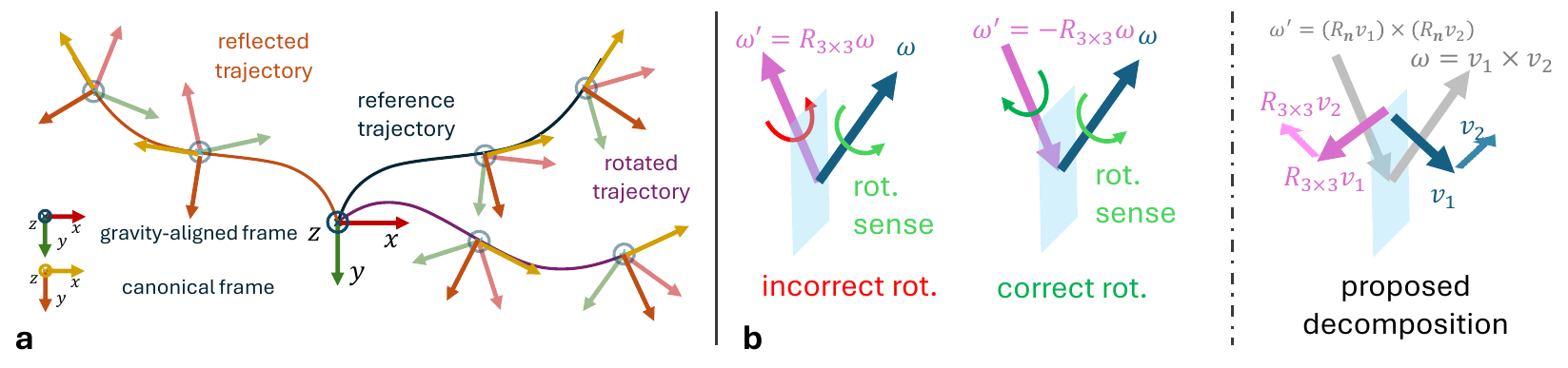}
    \vspace{-6mm}
    \caption{Symmetries in neural inertial odometry. (a) An IMU undergoes three trajectories in $xy$-plane, each related to a reference (blue) via rotation (purple) and/or reflection (orange) around gravity (parallel to the $z$-axis). At a fixed time, IMU measurements on different trajectories, expressed in the corresponding local gravity-aligned frame (red-green) differ only by an unknown yaw roto-reflection $R_{3\times3}$. Mapping these measurements to a canonical frame (yellow-red) that transforms equivariantly under roto-reflections of the trajectory eliminates this ambiguity enhancing the sample efficiency of downstream neural networks. (b) Expressed in alternative roto-reflected frames, acceleration and angular rates transform as $a'=R_{3\times3}a$ and $\omega'=\text{det}(R_{3\times3})R_{3\times3}\omega$. Angular rate must follow the right-hand-rule, and thus be also inverted when reflected. To ensure a similar transformation rule as $a$, we decompose $\omega=v_1\times v_2$ and process $v_1,v_2$ instead, which transform as $v_{1/2}'=R_{3\times3}v_{1/2}$}
    \label{fig:equivariance_illustration}
\vspace{-5.5mm}
\end{figure*}

% To reduce the data variability in IMU measurements, we map them to a \emph{gravity-aligned frame} (i.e. perform \emph{gravity-alignment}), which has its z-axis aligned with the gravity direction. We do this by simple rotations. 
We simplify the learning of informative priors by suitably canonicalizing the IMU measurements in two steps:  
First, we gravity-align IMU measurements by rotating them such that the z-axis of the IMU frame coincides with the z-axis of the world frame. We estimate gravity with the help of~\citep{liu2020tlio} by taking the orientation (rotation) estimate from the current EKF state (see App.~\ref{App:EKF}). Later we show empirically that our method is insensitive to noise originating from this estimation step. In what follows we thus assume accelerations and angular rates to be expressed in the gravity-aligned frame and illustrate these frames in Fig.~\ref{fig:equivariance_illustration} (a) for three rotated trajectories. %As observed The predicted displacement is expressed in the World-frame as it is used as a measurement update to the EKF which has it's states in the fixed World-frame as . Hence, any error in the state estimate will affect only the EKF rather than the NN.
% Gravity alignment means that the Z-axes of the IMU-frame and the world-frame overlap. The rotation between the IMU and the world is a state in the Kalman-filter and its current estimate is used for the Z-axis alignment. Indeed, a state estimate will have an error but this will affect the Kalman Filter rather than the neural displacement prediction which will be expressed back to the world frame in the KF update step.
%
% To reduce the data variability in IMU measurements, we map the IMU data to a \emph{gravity-aligned frame} by rotating it to a frame with a gravity-aligned z-axis using the orientation estimated from the current EKF state or known orientation.  
% Typically, the gravity direction can be estimated from an auxiliary \emph{visual-inertial} odometry (VIO) method or can be estimated from the dominant direction of accelerometer readings at rest during training. During test time, the orientation estimated from the current state of the EKF or known orientation is used for gravity-alignment. 
Gravity alignment reduces data variability by two degrees of freedom. However, this frame is not unique, since simply rotating it around z or reflecting it across planes parallel to z (applications of rotations or roto-reflections from the groups $SO_{g}(3)=\{R\in SO(3)\vert Rg=g\}$ and $O_{g}(3)=\{R\in O(3)\vert Rg=g\}$) result in new valid gravity-aligned frames. Hence, secondly, we predict a canonical yaw frame and map the IMU data into this frame, which we later show to be subequivariant to roto-reflections. In what follows we will restrict our discussion to the $O_{g}(3)$ case but note that, where not explicitly stated, this discussion carries over to $SO_{g}(3)$ as well. 
Next, we will introduce our canonicalization procedure to ensure better network generalization. 

\vspace{-2mm}
\section{Methodology}
\label{methodology}
\vspace{-3mm}
Our goal is to predict a canonical yaw frame $F=\Psi(\{(a_i, \omega_i)\}_{i=1}^n)\in O_{g}(3)$ from data, which generalizes across arbitrary yaw orientations. We use this frame to map IMU data into a canonical frame before giving as input to the neural network, and mapping the outputs back, (see Fig.~\ref{pipeline} (a)),\emph{i.e.}
{\small
\begin{align}
    d', \Sigma' &= \Phi(\{(a'_i,\omega'_i)\}_{i=1}^n) \quad\quad \text{ with } a'_i=\rho_a(F^{-1})a_i \text{ and } 
    \omega'_i=\rho_\omega(F^{-1})\omega_i,
\end{align}
}
where $a',\omega'$ are expressed in the canonical frame. Finally, we map $d',\Sigma'$ back to the original frame via $d=\rho_d(F )d'$ and $\text{vec}(\Sigma)=\rho_\Sigma(F )\text{vec}(\Sigma')$. Here $\text{vec}(.)$ stacks the columns of $\Sigma$ into a single vector, and $\rho$ is a homomorphism that maps group elements $F$ to corresponding matrices, called \emph{matrix representations}. These capture the transformation of $a,\omega,d$ and $\Sigma$ under the action of $F$.

While $\rho_a(F)=\rho_d(F)=F_{3\times 3}$, with $F_{3\times 3}\in\mathbb{R}^{3\times 3}$ being the rotation matrix corresponding to element $F$, covariances transform as $\rho_\Sigma(F) =F_{3\times 3}\otimes F_{3\times 3}$, where $\otimes$ is the Kronecker product. Unfortunately, reflections ($\text{det}(F_{3\times3})=-1$) induce a reflection \emph{and inversion} of angular rates to preserve the right-hand-rule (see Fig.~\ref{fig:equivariance_illustration} b), \emph{i.e.} $\rho_\omega(F)=\text{det}(F_{3\times3})F_{3\times3}$, and for reflections. As discussed later $\rho_\omega(F)\neq \rho_a(F)$ hinders joint processing of accelerations and angular rates. Next, we will discuss the design of $\Psi$ which ensures generalization across arbitrary yaw rotations.

%Hence, current equivariant networks that handle group representations cannot be used with $\omega$ and $a$ as input. If we map $\omega$ to a space where its components transform under same group representation as that of $a$ we can use a common equivariant network.
%We discuss next, how to decompose $\omega$ into components which do transform consistently with $a$.
\vspace{-2mm}
\subsection{Equivariant Frame}
\label{equivariant_frame}
\vspace{-3mm}
% This is equivalent to strict $O(2)$ equivariance if we treat the projection to the plane as the vector and projection to the gravity axis as invariant scalars
Here we derive a property of the frame network $\Psi$ such that it can generalize to arbitrary roto-reflections of the IMU body frame. To generalize, canonical IMU measurement inputs $a'_i,\omega'_i$ to the network must look identical under arbitrary roto-reflections $R\in O_{g}(3)$. Let $a_i,\omega_i$, and $a_i^*,\omega_i^*$  denote quantities before and after application of $R$. Then $a_i^*=\rho_a(R)a_i$ and $\omega_i^*=\rho_\omega(R)\omega_i$. Enforcing identical inputs under both rotations, \emph{i.e.} ${a^{*}}'_i=a'_i$ we have 
{\small
\begin{equation}
    \label{eq:invariant}
    {a^{*}}'_i=\rho_a({{F}^*}^{-1})\rho_a(R) a_i=a'_i=\rho_a({{F}}^{-1})a_i,
\end{equation}
}
and a similar equation for $\omega$. We see that choosing $F^*=R F$ satisfies this equality, leveraging the fact that $\rho_a$ is a homomorphism, \emph{i.e.} $\rho_a((RF)^{-1})=\rho_a(F^{-1}R^{-1})=\rho_a(F^{-1})\rho_a(R)^{-1}$. This equality puts a constraint on the neural network that estimates $F$, namely 
{\small
\begin{equation}
R F = \Psi(\{(\rho_a(R)a_i,\rho_\omega(R)\omega_i)\}_{i=1}^n)
\end{equation}
}
\emph{i.e.} $\Psi$ must be a function that is \emph{equivariant} with respect to group actions by elements from $O_{g}(3)$. Since this is a subgroup of $O(3)$ we also say that $\Psi$ must be \emph{subequivariant} with respect to $O(3)$. 

In addition, this equivariance property of $\Psi$ induces end-to-end equivariance to predicted displacements $d=\rho_d(F)d'$ and covariances $\text{vec}(\Sigma)=\rho_\Sigma(F)\text{vec}(\Sigma')$. This is because 
{\small
\begin{align}
    d^* &= \rho_d(F^*) {d^*}' = \rho_d(RF) d' = \rho_d(R) \rho_d(F)d' = \rho_d(R)d\\
    \text{vec}(\Sigma^*) &= \rho_\Sigma(F^*) {\text{vec}({\Sigma'}^*)} = \rho_\Sigma(RF) \text{vec}(\Sigma') = \rho_\Sigma(R) \rho_\Sigma(F)\text{vec}(\Sigma') = \rho_\Sigma(R)\text{vec}(\Sigma)
\end{align}
}
using, again, the homomorphism of $\rho$ and the fact that $d',\text{vec}(\Sigma')$ are, by construction of~\eqref{eq:invariant}, invariant to rotations by $R$.

\textbf{Diagonal Covariance: } We show empirically in Sec.~\ref{ablation_study_section} that the diagonal parameterization of $\Sigma'$ aids in stabilization  and convergence of the network. Therefore, we assume the displacement uncertainties $\Sigma_{d,xz}=\Sigma_{d,yz}=0$ and without loss in generality choose $\Sigma'=\text{diag}(e^{2u_x},e^{2u_y},e^{2u_z})$, where $u_x,u_y,u_z$ are learnable, as in TLIO~\citep{liu2020tlio}. Since our network predicts $\text{vec}(\Sigma)=\rho_\Sigma(F)\text{vec}(\Sigma')$, where both $\Sigma'$ and $F$ are learned, the resulting covariance is $\Sigma=F\Sigma'F^T$ (in matrix format). Via the transformation $F$ we can learn arbitrarily rotated $\Sigma$ in the $xy$-plane. We posit that this forces the frame network $\Psi$ to learn $F$ that aligns with the principle axes of the statistical uncertainty in displacement $\Sigma_d$. See App.~\ref{covariance_pre} for details on covariance parameterizations. Writing the singular value decomposition (SVD) we see that $\Sigma_d = U \text{diag}(\Sigma_{xx},\Sigma_{yy},\Sigma_{zz}) U^T$. By inspection, this uncertainty is matched when $F$ aligns with principle directions $U$ and $\Sigma'$ aligns with the true uncertainties in those directions.

% Since our network predicts $\text{vec}(\Sigma)=\rho_\Sigma(F)\text{vec}(\Sigma')$, where both $\Sigma'$ and $F$ are learned, we can leverage some redundancy, to simplify the parametrization of $\Sigma'$, without loss in generality. In particular, we choose $\Sigma'=\text{diag}(e^{2u_x},e^{2u_y},e^{2u_z})$, where $u_x,u_y,u_z$ are learnable, as in TLIO~\cite{liu2020tlio}, resulting in covariance $\Sigma=F\Sigma'F^T$ (where we now write it in matrix format). Via the transformation $F$ we can learn arbitrarily rotated $\Sigma$ in the $xy$-plane. We posit that this forces the frame network $\Psi$ to learn $F$ that align with the principle axes of the statistical uncertainty in displacement $\Sigma_d$, provided that the $z$-component of $d$ is independent of the $xy$-components, \emph{i.e.} $\sigma^2_{d,xz}=\sigma^2_{d,yz}=0$. See Appendix~\ref{covariance_pre} for details on covariance parameterizations. Writing the singular value decomposition (SVD) we see that $\Sigma_d = U \text{diag}(\sigma_x^2,\sigma_y^2,\sigma_z^2) U^T$. By inspection, we see that this uncertainty is matched when $F$ aligns with principle directions $U$ and $\Sigma'$ aligns with the true uncertainties in those directions. 

In the next section, let us now discuss the specific issue that arises when designing an equivariant frame network to process both $a_i$ and $\omega_i$, and the specific preprocessing step to remedy it. 
\vspace{-3mm}
\subsection{Decomposition of Angular Rates}
\vspace{-3mm}
\label{preprocessing}
As previously discussed $a$ and $\omega$ transform under different representations $\rho_a\neq\rho_\omega$. This hinders joint feature learning since this would entail forming linear combinations of $a$ and $\omega$, and these linear combinations will not transform under $\rho_a$ or $\rho_\omega$. We propose a preprocessing step that decomposes $\omega_i$ into perpendicular vectors $v_{1,i},v_{2,i}$ via a bijection $\mathcal{F}:\mathbb{R}^3\rightarrow \mathbb{R}^{3}\times \mathbb{R}^{3}$:
{\small
\begin{equation}
\mathcal{F}(\omega) = (v_1,v_2)= \left(\sqrt{\Vert \omega\Vert}\frac{w_1}{\Vert w_1\Vert}, \sqrt{\Vert \omega\Vert}\frac{w_2}{\Vert w_2\Vert}\right) \quad\quad \mathcal{F}^{-1}(v_1, v_2) = \omega = v_1 \times v_2
\end{equation}
}
We define $w_1=[-\omega_y\, \omega_x\, 0]^T$ and $w_2=\omega\times w_1$. %This mapping is a bijection since angular rates can be simply reconstructed via $\omega=v_{1}\times v_{2}$. 
If $\omega_{x}=\omega_y = 0$, we use $w_1=a \times \omega$ and if both $\omega_{x}=\omega_y = 0$ and $a \times \omega = \bm{0}$, we use $w_1=\omega \times [1\,0\,0]^T$.

Fig.~\ref{fig:equivariance_illustration} (b) shows that $v_{1}$ and $v_2$ transform with representation $\rho_{v_1}(F)=\rho_{v_2}(F)=F_{3\times 3}$. Let variables with $*$ denote transformed vectors according to rotation $R$. Their cross product has the desirable property $\omega^*=v^*_1\times v^*_2=(R_{3\times 3}v_{1})\times(R_{3\times 3}v_{2})=\text{det}(R_{3\times 3})R_{3\times 3}(v_1\times v_2)=\text{det}(R_{3\times 3})R_{3\times 3}\omega=\rho_{\omega}(R)\omega$, using the standard cross-product property $(Ax)\times (Ay) = \text{det}(A)A(x\times y)$, and recalling that $R_{3\times 3}$ is the matrix representation of $R$. The group action on $\omega$ exactly coincides with what was derived in Sec.~\ref{problem_setup}.
We only use this decomposition when with $O_{g}(3)$, where we process $a,v_1,v_2$ in a unified way. For $SO_{g}(3)$, we process $a,\omega$ which transform similarly since $\text{det}(R_{3\times 3 })=1$.
\vspace{-2pt}
\subsection{Transition to $O(2)$ Equivariance and Basic Network Layers}
\label{basic_layers}
\vspace{-3mm}
Expressed in the gravity-aligned frame, representations $R_{3\times 3}$ of $R\in O_{g}$ leave the z-axis unchanged, and can thus be decomposed into $R_{3\times 3}=R_{2\times 2}\oplus 1$, where the direct sum $\oplus$ constructs a block-diagonal matrix of its arguments. This decomposition motivates the decomposition $a_i = a_{i,xy}\oplus a_{i,z}$ and $v_{1/2,i}=v_{1/2,i,xy}\oplus v_{1/2,i,z}$, where $\oplus$ concatenates the $xy$-coordinates of each vector which transform with representation $R_{2\times 2}$ and its $z$ component which transforms with representation $1$, \emph{i.e.} $z$ is invariant. This means that the $xy$-components transform according to representations of group $O(2)$ and implies that $O(2)\cong O_{g}(3)$. 
Inspired by~\cite{villar2021scalars}, we design our frame network to learn universally $O(2)$ equivariant outputs from invariant features alongside 2D vector features.

We convert the sequence of $n$ IMU measurements into $n\times C_0^s$ rotation invariant scalar features and $n\times 2\times C_0^v$ vector features. As vector features we select the $xy$-components of each input vector ($C_0^v=2$ for $SO(2)$ corresponding with $a_{i,xy},\omega_{i,xy}$ and $C_0^v=3$ for $O(2)$ corresponding with $a_{i,xy},v_{1,i,xy},v_{2,i,xy})$. Instead, as scalar features we select \emph{(i)} the $z$-components of each vector, \emph{(ii)} the norm of the $xy$-components of each vector, and \emph{(iii)} the pairwise dot-product of the $xy$-components of each vector. 
For $SO(2)$ we have $C_0^s=2 + 2 + 1=5$, while for $O(2)$ we have $C_0^s=3 + 3 + 3=9$.

While we process scalar features with multilayer perceptrons and standard 1-D convolutions, we process vector features with specific linear and convolution layers, and combine scalar and vector features with specialized non-linear layers.
\label{basic_layers}
\vspace{-2mm}
\paragraph{Equivariant Linear Layer} Following~\citet{villar2021scalars}, we design a $2D$ version of vector neuron~\citep{deng2021vector} to process the vector features, enhancing efficiency. Following~\citet{finzi2021practical}, we consider learnable linear mappings $v^{\text{out}}=Wv^{\text{in}}$, with input and output vector features  $v^{\text{in}},v^{\text{out}}\in\mathbb{R}^2$ and seek a basis of weights $W \in \mathbb{R}^{2\times2}$, which satisfy $R_{2\times 2}Wv=WR_{2\times 2}v$, i.e., equivariantly transform vector features $v\in\mathbb{R}^2$. This relation yields the constraint
{\small
\begin{align}
&(R_{2\times 2}\otimes R_{2\times 2}) \text{vec}(W) = \text{vec}(W),
\label{linear_cons}
\end{align}
}
Solving the above equation amounts to finding the eigenspace of the left-most matrix with eigenvalue 1. Such analysis for $R_{2\times2}\in SO(2)$ yields $W_{SO(2)}=w_1 I_{2\times 2} + w_2 R_{90}$, where $R_{90}$ denotes a $90$ degree counter-clockwise rotation in 2D, and $w_1,w_2\in\mathbb{R}$ are learnable weights. Similarly, for $O(2)$ we find  $W_{O(2)}=w_1 I_{2\times 2}$. 
Vectorizing this linear mapping to multiple input and output vector features we have 
the following $SO(2)$ and $O(2)$ equivariant linear layers: 
{\small
\begin{align}
    SO(2):\quad v^\text{out} = v^\text{in}W_{1}  + R_{90} v^\text{in} W_2\quad\quad
    O(2):\quad v^\text{out} = v^\text{in}W_{1} 
\end{align}
}
with $v^\text{in}\in \mathbb{R}^{2\times C_\text{in}}$, $v^\text{out}\in \mathbb{R}^{2\times C_\text{out}}$ and $W_1,W_2\in \mathbb{R}^{C_\text{in}\times C_{\text{out}}}$.  Note that the $SO(2)$ layer has twice as many parameters as the $O(2)$ layer.

%, given multi-channel input vector features $v^{in} \in \mathbb{R}^{2 \times C_{in}}$, for $O(2)$, our linear layer is: 
%\begin{align}
%    v^{out} = v^{in}W,
%\end{align}
%where $W \in \mathbb{R}^{C_{in} \times C_{out}}$;
%for $SO(2)$, our linear layer becomes:
%\begin{align}
%    v^{out} = v^{in}W_1+(v^{in})^oW_2,
%\end{align}
%where $W_1, W_2 \in\mathbb{R}^{C_{in} \times C_{out}}$. 
%\input{preprocess}
%In addition to $2D$ vector features, we also utilize invariant scalar features. For these invariant features, we employ a conventional linear layer, which doesn't break the subequivariance. 
We use the above linear components to design equivariant 1-D convolution layers by stacking multiple such weights into a kernel. Since the IMU data forms a time sequence, we implement convolutions across time. We visualize our Linear and Convolutional Layers in Fig.~\ref{pipeline}b for better understanding. 
\vspace{-2mm}
\paragraph{Nonlinear Layer} Previous works~\citep{weiler20183d, weiler2019general} propose various nonlinearities such as norm-nonlinearity, tensor-product nonlinearity, and gated nonlinearity for $SO(3)$ and $O(2)$ equivariance in an equivariant convolutional way; while~\citet{deng2021vector} applies per-point nonlinearity for vector features only. Since we already apply convolutions over time we simply apply a non-linearity pointwise. Unlike~\citet{deng2021vector}, we need to mix scalar and vector features and thus adapt the gated nonlinearity~\citep{weiler20183d} to pointwise nonlinearity.  
Specifically, for $n$ vector and scalar features $v^{\text{in}}\in \mathbb{R}^{n\times 2\times C},s^{\text{in}}\in \mathbb{R}^{n\times C}$, we concatenate the norm features $\Vert v^{\text{in}}\Vert \in \mathbb{R}^{n\times C}$ with $s^{\text{in}}$. We run a single MLP with an output of size $n\times 2C$, and split it into new norm features $\gamma\in \mathbb{R}^{n\times C}$ and new activations $\beta\in\mathbb{R}^{n\times C}$ which we modulate with a non-linearity $s^{\text{out}}=\sigma(\beta)$. Finally, we rescale the original vector features according to the new norm:
{\small
\begin{align}
    \gamma, \beta = \text{mlp}(\Vert v^{\text{in}}\Vert\oplus_c s^{\text{in}})\quad\quad
    v^{\text{out}} = \gamma v^{\text{in}}\quad\quad
    s^{\text{out}} = \sigma(\beta)
\end{align}
}
where $\oplus_c$ concatenates along the feature dimension.
%to generate new vector norm features then   scalars, i.e., we get the new hidden features $(v^{in}, s'^{in})$ with $s'^{in} = (\oplus_c norm(v_c^{in}))\oplus s^{in}$, where $c$ denotes the index of channel. 
%Subsequently, we only apply a conventional linear layer and activation on $s'^{in}$ to get $(\oplus_c s''_c)\oplus s^{out}$, and following we do multiplication of scalar and vector (i.e. tensor product) to get output feature $(v^{out}, s^{out})$, where $v^{out}=\oplus_c (s''_c v^{in}_c)$. 
See Fig.~\ref{pipeline}b for more details. 

\vspace{-3mm}
\section{Experiments}
\vspace{-3mm}
\label{datasets}
% \subsection{Datasets and Implementation}
To demonstrate the effectiveness of canonicalizing the IMU data using the equivariant framework, we apply it to two types of neural inertial navigation systems: 1) an end-to-end deep learning approach (RONIN), and 2) a filter-based approach with a learned prior (TLIO). 
Both architectures take IMU samples in a gravity-aligned frame without gravity compensation as input to the neural network, i.e., without removing the gravity vector from the accelerometer reading. While RONIN regresses only a 2D velocity, TLIO estimates the orientation, position, velocity, and IMU biases using an EKF. This EKF propagates states using raw IMU measurements and applies measurement updates with predicted displacement and uncertainty from a neural network. Sec.~\ref{ablation_study_section} presents extensive ablations on the various design choices for the subequivariant framework.  

\textbf{Datasets: }
Our TLIO variant is trained on the TLIO Dataset~\citep{liu2020tlio} and tested on TLIO and Aria Everyday Activities (Aria) Datasets~\citep{lv2024aria}. Our RONIN variant is trained on RONIN Dataset~\citep{herath2020ronin}. We train on only 50\% of the dataset that is open-sourced. We test our RONIN variant on three popular pedestrian datasets RONIN~\citep{herath2020ronin}, RIDI~\citep{Yan_2018_ECCV} and OxIOD~\citep{OXOID}, which specifically target 2D trajectory tracking. We use specific test datasets only to provide a fair comparison with prior work. See App.\ref{dataset_details} for more dataset details, and see App.\ref{network_details} and Fig.~\ref{pipeline} for more details and visualizations of the equivariant network. 

\textbf{Baselines: } We compare our method with TLIO with yaw augmentation~\citep{liu2020tlio}, on the 3D trajectory tracking benchmarks, as well as RONIN and RIO~\citep{cao2022rio} on the 2D trajectory benchmarks. We also report from~\citet{herath2020ronin} naive double integration (NDI) of IMU measurements, and RONIN results where authors have trained on 100\% of the RONIN training dataset. All other methods use only 50\% of this dataset. RIO extends RONIN with two additional features, denoted with $+J$ and $+TTT$. The $+J$ denotes joint optimization of an MSE loss on velocity predictions and cosine similarities with an equivariance constraint modeled using an auxiliary loss, while $+TTT$ adapts RONIN at test time using an Adaptive Test-Time-Training strategy. Finally, $+J+TTT$ combines both. 

% TLIO dataset is a headset dataset that consists of IMU raw data at 1kHz and ground truth obtained from MSCKF at 200 Hz for 200 sequences totaling 60 hours. We use their data splits for training, validation and testing. Aria~\cite{lv2024aria} is an open-sourced egocentric dataset that is collected using Project Aria Glasses. There are two IMUs on the left and right side of the headset of frequencies 800 and 1kHz respectively. They have two sources of ground truth- open and closed loop trajectory at 1kHz. Open loop trajectory is strictly causal while closed loop jointly processes multiple recordings to place them in a common coordinate system. This dataset consists of 143 recordings accumulating to 7.3 hrs which we use as a test dataset. The raw right IMU data is used to compare closed-loop trajectory with EKF results. The data was downsampled to 200Hz and preprocessed using the closed-loop trajectory to test the Neural Network trained on TLIO. For RONIN model, we use the two popular pedestrian datasets \textbf{RONIN}\cite{herath2020ronin} and \textbf{RIDI}\cite{Yan_2018_ECCV}. They both have IMU frequency 200 Hz. We use RONIN data splits to train and test their model with and without our framework. We report test results on both RIDI test and cross-subject datasets. RIDI results are presented after pre-processing the predicted trajectory with the Umeyama algorithm~\cite{umeyama1991} for fair comparison against other methods.
% \subsection{Experiment Results}
% \subsection{Baseline}
\textbf{Metrics: } The Neural Network performance (indicated with $*$) is evaluated using three metrics: Mean Squared Error (MSE) in $10 ^{-2}m^2$, Absolute Translation Error (ATE) in $m$, and Relative Translation Error (RTE) in $m$. For evaluating only the neural network, the trajectory is reconstructed via cumulative summation from the initial state as done in prior work. In case of TLIO architecture, we additionally evaluate the overall performance after integrating the neural network with the EKF and report the ATE in $m$, RTE in $m$, and Absolute Yaw Error (AYE) in degrees. Details of these metrics are provided in the App.~\ref{eval_metrics}. In what follows, we abbreviate \ours{} with +Eq. Frame.

\vspace{-3mm}
\subsection{Results using the TLIO Architecture}
\vspace{-3mm}

\begin{table}[t!]
\newcommand{\first}{\cellcolor{red!40}}
\newcommand{\second}{\cellcolor{orange!40}}
\newcommand{\third}{\cellcolor{yellow!40}}
\centering
% \begin{tabular}{lcccccccccc}
\begin{small}
\resizebox{\linewidth}{!}{
\begin{tabular}{lcccccccccccc}
\toprule
& \multicolumn{6}{c}{TLIO Dataset} & \multicolumn{6}{c}{Aria Dataset} \\
\cmidrule(lr){2-7} \cmidrule(lr){8-13}
Model & MSE*
& ATE & ATE* & RTE & RTE* &AYE &  MSE* & ATE & ATE* & RTE & RTE* &AYE  \\
&($10^{-2}m^2$)&($m$)&($m$)&($m$)&($m$)&(deg)&($10^{-2}m^2$)&($m$)&($m$)&($m$)&($m$)&(deg)\\
\midrule

TLIO&     3.333 & 1.722 &3.079 & 0.521 & 0.542 & \first{2.366} & 15.248 & 1.969 & 4.560 & 0.834 & 0.977 & 2.309\\
+ rot. aug.         & 3.242 & 1.812 &3.722 & 0.500 & 0.551 & \second{2.376} & 5.322 & 1.285 & 2.103 & 0.464 & 0.521 & \second{2.073}\\
\hline
+ \textbf{$\mathbf{SO(2)}$ Eq. Frame } &\second{3.194} & \second{1.480} &\second{2.401} & \second{0.490} & \second{0.501} & 2.428 & \second{2.457} & \second{1.178} & \second{1.864} & \second{0.449} & \second{0.484} & 2.084 \\
+ \textbf{$\mathbf{O(2)}$ Eq. Frame }   & \first{2.982} & \first{1.433} &\first{2.382} & \first{0.458}& \first{0.479} & 2.389& \first{2.304} & \first{1.118} &\first{1.850} & \first{0.416} & \first{0.465} & \first{2.059}\\
\bottomrule
\end{tabular}}
\vspace{-3mm}
\caption{ Trajectory errors without (labelled with *) and with EKF, lower being better. + rot. aug. is trained with yaw augmentations. Lowest, and second lowest values are marked in red and orange, and our methods are in bold.}
\label{nn_tlio_table}
\end{small}
\vspace{-5mm}
\end{table}

\begin{figure}[t!]
\begin{center}
\centerline{\includegraphics[width=\linewidth]{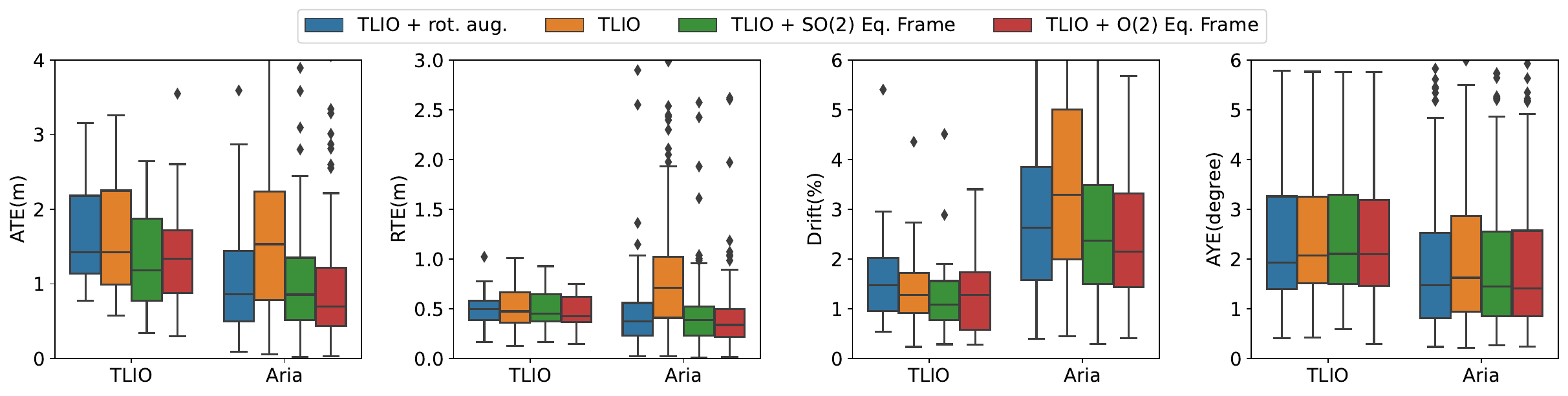}}
\vspace{-3mm}
\caption{Trajectory errors for EqNIO applied to TLIO compared to vanilla TLIO trained with and without yaw augmentations on TLIO and Aria Datasets visualized with a box plot. }
\label{boxplot_ekf_tlio}
\end{center}
\vspace{-5mm}
\end{figure}
\begin{figure}[t!]
\begin{center}
\centerline{\includegraphics[width=0.9\linewidth]{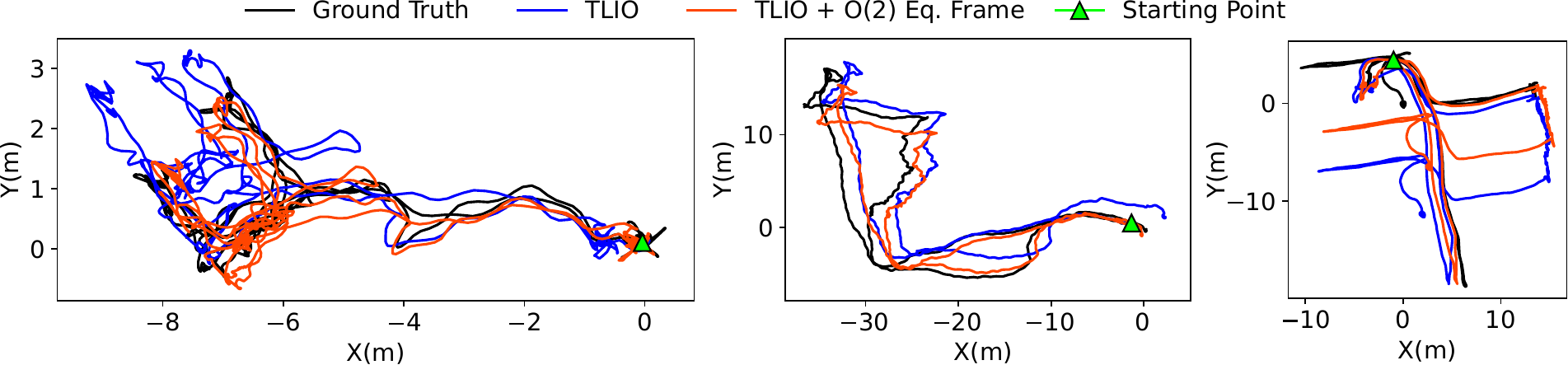}}
\vspace{-3mm}
\caption{Groundtruth (black), and predicted trajectories on the TLIO Dataset by baseline TLIO (Blue), our best method applied to TLIO (Red). Left and right are difficult trajectories, while the middle trajectory has medium difficulty.}
\label{traj_tlio_main}
\end{center}
\vspace{-8mm}
\end{figure}
Tab.~\ref{nn_tlio_table} compares baseline TLIO, trained with yaw augmentations as in~\citep{liu2020tlio} (+ rot.aug.), TLIO without augmentation (termed TLIO), and our two methods applied to TLIO without yaw augmentations (termed +$SO(2)$ and +$O(2)$ Eq. Frame). As seen in Tab.~\ref{nn_tlio_table}, +$O(2)$ Eq. Frame outperforms TLIO on metrics that ignore the EKF (with *) by a large margin of 57\%, 12\%, and 11\%  on MSE*, ATE*, and RTE* respectively. The +$SO(2)$ Eq. Frame model follows closely with 54\%, 11\% and 7\% respectively on the Aria Dataset. The performance of our methods is consistent across TLIO and Aria Datasets illustrating our generalization ability. Tab.~\ref{nn_tlio_table} and Fig.~\ref{boxplot_ekf_tlio} shows our method surpasses the baseline on most metrics while remaining comparable in AYE. The superior performance of our model as compared to baseline TLIO when the NN is combined with EKF (\emph{i.e.}, performance on ATE, RTE and AYE metrics) is attributed to its generalization ability when the orientation estimate is not very accurate as well as the equivariant covariance predicted by the network. 
% The AYE of our method is comparable to TLIO with a difference of magnitude 0.02 degrees and the equivariant network is more tolerant to yaw perturbations.
See Fig.~\ref{traj_tlio_main} and App.~\ref{visualization_tlio_main} for trajectory plots from the TLIO test dataset.
 \vspace{-3mm}
\subsection{Results using the RONIN Architecture}
\begin{table}[t!]
\newcommand{\first}{\cellcolor{red!40}}
\newcommand{\second}{\cellcolor{orange!40}}
\newcommand{\third}{\cellcolor{yellow!40}}
\centering
\begin{small}
\resizebox{\linewidth}{!}{
\begin{tabular}{lcccccccccc}
\toprule
& \multicolumn{2}{c}{RONIN-U} & \multicolumn{2}{c}{RONIN-S}&\multicolumn{2}{c}{RIDI-T}&\multicolumn{2}{c}{RIDI-C}& \multicolumn{2}{c}{OxIOD} \\
\cmidrule(lr){2-3} \cmidrule(lr){4-5}\cmidrule(lr){6-7}\cmidrule(lr){8-9}\cmidrule(lr){10-11}
Model & ATE* & RTE* & ATE* & RTE*& ATE* & RTE* &ATE* & RTE*&ATE* & RTE* \\
(RONIN) & (m)&(m)& (m)&(m)& (m)&(m)& (m)&(m)& (m)&(m)\\
\midrule
+ 100\% data         & 5.14 & 4.37 & 3.54  &2.67 & 1.63 & 1.91 & 1.67 & 1.62 & 3.46 & 4.39 \\
+ 50\% data $\dagger$        & 5.57 & 4.38 & - & - & 1.19 & 1.75 & - & - & 3.52 & 4.42 \\
+ 50\% data + J $\dagger$         & \second{5.02} & 4.23 & - & - & 1.13 & 1.65 & - & - & 3.59 & 4.43\\
+ 50\% data + TTT $\dagger$     & 5.05 & \second{4.14} & - & - & 1.04 & 1.53 & - & - & 2.92& 3.67\\
+ 50\% data + J +TTT $\dagger$     & 5.07 & 4.17 & - & - & 1.03 &\first{1.51} & - & -&2.96&3.74\\
\hline
+ 50\% data + \textbf{$\mathbf{SO(2)}$ Eq. Frame}     & 5.18 & 4.35 & 3.67 & 2.72 & \second{0.86} & 1.59 & \first{0.63} & \first{1.39} & \first{1.22} & \second{2.39}\\
+ 50\% data + \textbf{$\mathbf{O(2)}$ Eq. Frame} & \first{4.42} & \first{3.95} & \first{3.32} & \first{2.66} &\first{0.82} & \second{1.52} &\second{0.70} & \second{1.41} & \second{1.28} &\first{2.10} \\
\hline
Naive Double Integration (NDI)&458.06&117.06&675.21&1.6948&31.06&37.53&32.01&38.04&1941.41&848.55\\
\bottomrule
\end{tabular}}
\vspace{-3mm}
\caption{Trajectory errors (to two decimals, as ~\cite{herath2020ronin}) with the RONIN architecture (lower is better), on the RONIN Unseen (-U), RONIN Seen (-S), RIDI Test (-T), RIDI Cross Subject (-C), and the Oxford Inertial Odometry Datasets (OxIOD). Lowest and second lowest values are red and orange respectively, and our method is in bold. Results with $\dagger$ are from~\cite{cao2022rio}.}
\label{ronin_results}
\end{small}
\vspace{-4mm}
\end{table}
 \vspace{-3mm}
%Table~\ref{ronin_results} compares RONIN~\cite{herath2020ronin}, RIO~\cite{cao2022rio} and our methods. The RONIN and RIO results are as reported in their paper. As the complete RONIN data has not been made available, we use only the published 50\% dataset to train RONIN using our method. RIO reports the RONIN original model trained on this dataset as B-ResNet. J-ResNet is RONIN trained to jointly optimize the MSE loss on velocity prediction and Cosine Similarity imposing the equivariance constraint. RIO also reports the results of applying their novel Adaptive Test-Time-Training strategy on these two models as B-ResNet-TTT and J-ResNet-TTT respectively. They do not report results on RONIN Seen (RONIN-S) and RIDI Cross Subject (RIDI-C). We can see from Table~\ref{ronin_results} that the equivariant methods surpass even the original model trained on double the amount of data by a large margin. The RONIN paper claims convergence of the model above 100 epochs, while our Eq Frame O(2) model converges at 38th epoch. These results demonstrate the superior generalization ability of fully equivariant architecture over approximate equivariance achieved via augmentations in RONIN and an auxiliary equivariance loss in RIO. The computational and memory costs are also higher for RIO as compared to our methods as they process the data 5 times (1 original and 4 discrete rotated versions) to the model at test time. They also build a deep ensemble to estimate covariance and store 2 sets of models as part of their TTT strategy.
On the RONIN dataset we compare against RONIN and RIO$\dagger$ (indicated with +J, +TTT, and +J+TTT). RIO does not provide results for RONIN Seen Dataset (RONIN-S) or RIDI Cross Subject Dataset (RIDI-C). As the RONIN base model does not use an EKF, we only report metrics with $*$. As Seen in Tab.~\ref{ronin_results}, our methods significantly outperform the original RONIN by a large margin of 14\% and 9\% on ATE* and RTE* respectively even on the RONIN-U dataset. Our methods have better generalization as seen on RIDI-T and OxIOD, outperforming even +J+TTT$\dagger$ by a margin of 56\% and 43\% on ATE* and RTE* respectively on OxIOD Dataset. The +$O(2)$ Eq. Frame model converges at 38 epochs compared to over 100 in RONIN implying faster network convergence with our framework as compared to data augmentation. This demonstrates superior generalization of our strictly equivariant architecture. RIO's approach, involving multiple data rotations, test optimization, and deep ensemble at test time, would result in higher computational and memory costs as compared to our method.
Finally, the comparison with NDI highlights the need for a neural displacement prior.

\vspace{-4mm}
\section{Ablation Study}
\label{ablation_study_section}
\vspace{-3mm}

\begin{table}[t!]
\newcommand{\first}{\cellcolor{red!40}}
\newcommand{\second}{\cellcolor{orange!40}}
\newcommand{\third}{\cellcolor{yellow!40}}
\centering
\begin{small}
\resizebox{\linewidth}{!}{
\begin{tabular}{lcccccccccccc}%{p{2.1cm}p{.54cm}p{.54cm}p{.54cm}p{.54cm}p{.54cm}p{.54cm}p{.55cm}p{.54cm}p{.54cm}p{.54cm} p{.54cm} p{.54cm}}
\toprule
& \multicolumn{6}{c}{TLIO Dataset} & \multicolumn{6}{c}{Aria Dataset} \\
\cmidrule(lr){2-7} \cmidrule(lr){8-13}
Model & MSE*
& ATE & ATE* & RTE & RTE* &AYE &  MSE* & ATE & ATE* & RTE & RTE* &AYE  \\
&($10^{-2}m^2$)&($m$)&($m$)&($m$)&($m$)&(deg)&($10^{-2}m^2$)&($m$)&($m$)&($m$)&($m$)&(deg)\\
\midrule
TLIO  & 3.333         &1.722        & 3.079            & 0.521           & 0.542          & \first{2.366} & 15.248         &1.969         & 4.560         &0.834         & 0.977             &2.309\\
+ rot. aug.          & 3.242         &1.812        & 3.722            & 0.500           & 0.551          & \second{2.376}& 5.322          &1.285         & 2.102         &0.464         & 0.521             &\third{2.073}\\
+ rot. aug. + more layers   & 3.047         &1.613        & 2.766            & 0.524           & 0.519          & 2.397         & 2.403          &1.189         & 2.541         &0.472         & 0.540             &2.081\\
+ rot. aug. + Non Eq. Frame      &\third{3.008}  &\first{1.429}& 2.443            & 0.495           & 0.496          & 2.411         & 2.437          &1.213         & 2.071         &0.458         & 0.508             &2.096\\
+ rot. aug. + PCA Frame& 3.473         &1.506        & 2.709            & 0.523           & 0.535          & 2.459         & 6.558          &1.717         & 4.635         &0.771         & 0.976             &2.232\\
+ $SO(2)$ Eq. Frame + S& 3.331         &1.626        & 2.796            & 0.524           & 0.536          & 2.440         & 2.591          &\second{1.146}& 2.067         &0.466         & 0.517             &2.089\\
+ $SO(2)$ Eq. Frame + P & 3.298         &1.842        & 2.652            & 0.588           & 0.523          & 2.537         & 2.635          &1.592         & 2.303         &0.585         & 0.539             &2.232\\
+ \textbf{$\mathbf{SO(2)}$ Eq. Frame }   & 3.194         &\third{1.480}& \third{2.401}   & \third{0.490}   & 0.501          & 2.428         & 2.457          &1.178         & 1.864         &\third{0.449} &0.484              &2.084\\
+ $O(2)$ Eq. Frame + S & 3.061         &1.484        & 2.474            & \second{0.462}  &\third{0.481}   & 2.390         &\third{2.421}   &\third{1.175}&\first{1.804}   &\second{0.421}&\first{0.458}      &\first{2.043}\\
+ $O(2)$ Eq. Frame + P  &\second{2.990} &1.827        &\first{2.316}     & 0.578           &\first{0.478}  & 2.534         &\second{2.373}  &1.755         & \third{1.859} &0.564         & \third{0.468}     &2.223\\
+ \textbf{$\mathbf{O(2)}$ Eq. Frame }&\first{2.982}  &\second{1.433}& \second{2.382}   & \first{0.458}   &\second{0.479}   & \third{2.389} &\first{2.304}   &\first{1.118} &\second{1.849} &\first{0.416} & \second{0.465}    &\second{2.059}\\
Eq CNN        & 3.194         &1.580        & 3.385            & 0.564           & 0.610          & 2.394         & 8.946          &3.223         & 6.916         &1.091         & 1.251             &2.299\\
\bottomrule
\vspace{-3.5mm}
\end{tabular}}
\caption{Ablations of our model (bold) with the TLIO architecture, lower is better. %Neural network ablation studies are maked with (NN). MSE is multiplied by 100.
We test non-equivariant frames (+Non Eq. Frame), PCA-based frames (+PCA Frame), $SO(2)$ equivariant frames (+$SO(2)$ Eq. Frame), and $O(2)$ equivariant frames (+$O(2)$ Eq. Frame), and the effect of training with yaw augmentations (+ rot. aug.). We also test $xy$-isotropic (+S) and Pearson-based covariance parameterizations. Eq CNN is a fully equivariant CNN. Red, orange and yellow indicates the first, second and third lowest values respectively.}
\label{NN_ablation_table}
\end{small}
\vspace{-7mm}
\end{table}
Here, we show the necessity of incorporating equivariance in inertial odometry, the choice of equivariant architecture and covariance, and finally sensitivity analysis to estimated gravity direction. We present all the ablations using the TLIO base model in Tab.~\ref{NN_ablation_table}, both with and without integrating the EKF. App.~\ref{aug_test_results} further contains the performance of all models above on a test dataset which is augmented with rotations and/or reflections. App.~\ref{sequence_length_ablation} and App.~\ref{gravity_direction_ablation_section} present a sensitivity study on the input sequence length and estimated gravity direction.
% What ablation and why we do it?
% why do we think we achieve better on that ablation, achieve better performance proves what

%\subsection{Baseline Ablation}
\textit{\textbf{Baseline Ablation: }}\textbf{Is yaw augmentation needed when the input is in a local gravity-aligned frame?}
%\paragraph{Is yaw augmentation needed when the input is in a local gravity-aligned frame?} %We trained TLIO with the same hyperparameters without the yaw augmentation during training. The results for TLIO and TLIO wo augmentation presented in both the tables show that augmentation improves the generalization ability of the network which leads to better results on all metrics. These results further motivate the need for equivariance to guarantee the generalization of the network.
We trained TLIO both with and without yaw augmentation using identical hyperparameters and the results in Tab.~\ref{NN_ablation_table} (rows 1 and 2) reveal that augmentation enhances the network's generalization, improving all metrics for the Aria dataset with the lowest margin of 10\% on AYE and highest margin of 65\% for MSE*. This underscores the importance of equivariance for network generalization.
%\paragraph{Does a Deeper TLIO with comparable number of parameters match the performance of equivariant methods?} %The original TLIO neural architecture has a residual depth of [2, 2, 2, 2] which we increase to [3,3,3,3] inorder to be comparable with our equivariant method applied to TLIO. Our Eq Frame O(2) model has fewer parameters owing to the removal of orthogonal basis in vector neuron-based architecture of SO(2). The results in Table~\ref{NN_ablation_table} and~\ref{EKF_ablation_table} for Deeper TLIO when compared to our methods demonstrates that a bigger non-equivariant network with augmentations does not achieve exact equivariance needed for inertial odometry.
\textbf{Does a Deeper TLIO with a comparable number of parameters match the performance of equivariant methods?} We enhanced the residual depth of the original TLIO architecture from 4 residual blocks of depth 2 each to depth 3 each (row 3) to match the number of parameters with our $+SO(2)$ Eq. Frame model (row 8). Despite having fewer parameters due to the removal of the orthogonal basis in $SO(2)$ vector neuron-based architecture, +$O(2)$ Eq. Frame model (row 11) still outperformed the deeper TLIO. The data from Tab.~\ref{NN_ablation_table} demonstrate that merely increasing network's size, without integrating true equivariance, is insufficient for achieving precise inertial odometry.
\begin{figure}[t!]
\vspace{-4mm}
\begin{center}
\centerline{\includegraphics[width=1.2\linewidth]{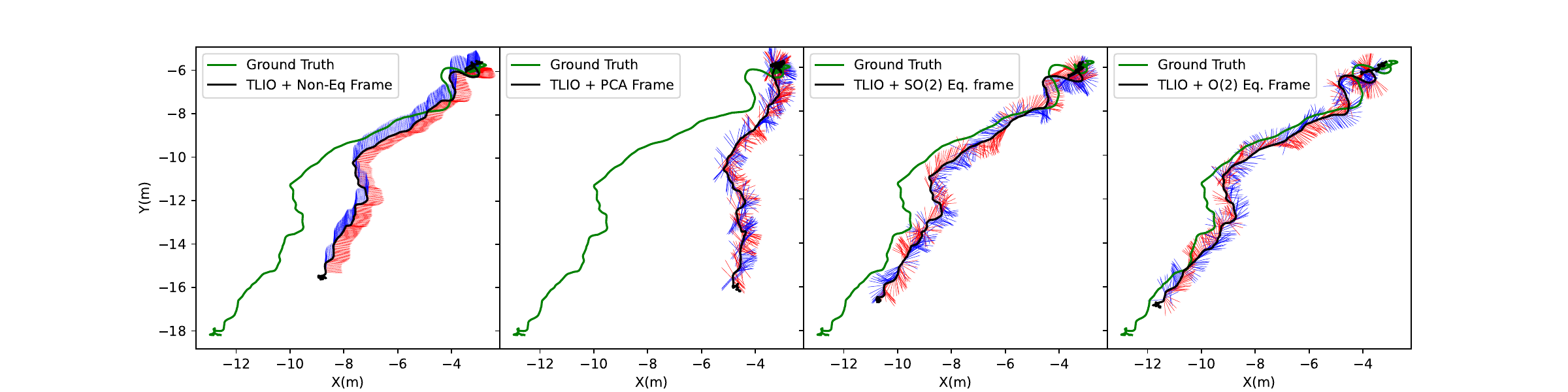}}
%\vspace{-1.8ex}
\vspace{-4mm}
\caption{Integrated trajectories on the Aria Dataset with ground truth (green) and predicted trajectories (black), as well as the equivariant frame basis vectors (blue/red).}
\vspace{-10mm}
\label{frames_aria}
\end{center}
\end{figure}
%\subsection{Frame Ablation}
%\paragraph{Can a non-equivariant MLP predict meaningful frames?} We train TLIO with augmentation and the same hyperparameters and an additional MLP with the same architecture as the our method to predict a frame. We observe that Non-Eq Frame + TLIO overfits to the training distribution and the predicted frame is not meaningful as seen in Figure~\ref{frames_aria}.
%\paragraph{Can frames predicted using PCA achieve the same performance as equivariant methods?} PCA frame does not generalise to Aria and has worse performance as compared to original TLIO on Aria dataset. We think this is because PCA is not smooth and is easily affected by noise as seen in Figure~\ref{frames_aria}. PCA cannot differentiate between SO(2) and O(2). From the Figure~\ref{frames_aria}, it can be seen that O(2) does not have frames as smooth as SO(2) as the reflected bends have reflected frames.

\textit{\textbf{Frame Ablation: }}\textbf{Can a non-equivariant MLP predict meaningful frames?} We trained TLIO with yaw augmentation and identical hyperparameters alongside an additional MLP mirroring the architecture of our method to predict a frame and term this baseline +Non Eq. Frame (row 4). We observed that +Non Eq. Frame tends to overfit to the TLIO dataset, and thus produce worse results on the Aria dataset. The predicted frames also poorly correlate with the underlying trajectory, as illustrated in Fig.~\ref{frames_aria}.
%We train TLIO with augmentation and the same hyperparameters and an additional MLP with the same architecture as the our method to predict a frame. We observe that Non-Eq Frame + TLIO overfits to the training distribution and the predicted frame is not meaningful as seen in Figure~\ref{frames_aria}.
\textbf{Can frames predicted using PCA (handcrafted equivariant frame) achieve the same performance?} Using PCA to generate frames leads to underperformance on the Aria dataset, and worse results than the original TLIO which is likely due to PCA's noise sensitivity as shown in Fig.~\ref{frames_aria}.  Additionally, PCA cannot distinguish between $SO(2)$ and $O(2)$ transformations.  Fig.~\ref{frames_aria} also shows that $O(2)$ does not have frames as smooth as $SO(2)$ as the reflected bends have reflected frames.
\vspace{-2mm}
\begin{wrapfigure}{r}{0.3\textwidth}
  \centering
  \vspace{-3mm}
  \includegraphics[width=0.25\textwidth]{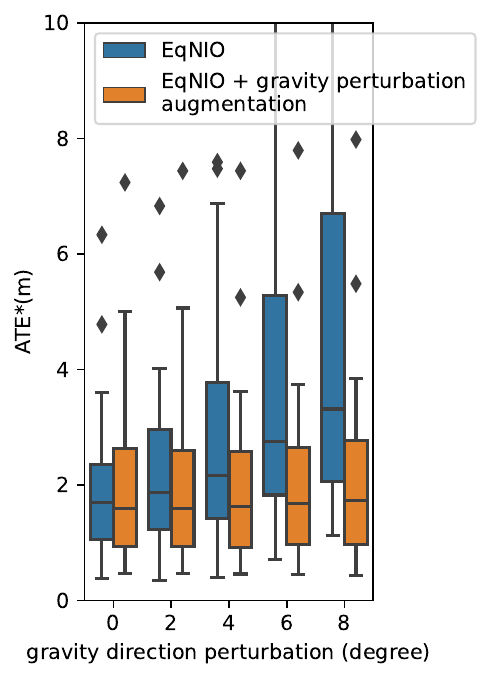}
  \vspace{-3mm}
  \caption{Sensitivity to gravity axis perturbation, and effectiveness of perturbation augmentation during training.}
\label{ate_box_grav}
\vspace{-8mm}
\end{wrapfigure}

\textit{\textbf{Architecture Ablation: }}\textbf{Does a fully equivariant architecture outperform our frame-based approach ?} We trained a fully equivariant 1-D CNN using the layers in Sec.~\ref{basic_layers}. Tab.~\ref{NN_ablation_table} (row 12) shows our frame-based methods outperforming the equivariant CNNs, likely by leveraging the power of scalars and conventional backbones. We believe the fully equivariant architecture is overly restrictive, while also requiring a full network redesign. By contrast, our method can flexibly integrate existing state-of-the-art displacement priors.
% \begin{itemize}
%     \item eqcnn -  having full equivariant network constrains the model too much
%     \item leveraging power of scalars are universal
%     \item frame more efficient and effective
%     \item leverage of conventional current backbone
%     \item the frame also makes it very easy to adopt into any neural inertial navigation system
% \end{itemize}

%\subsection{Covariance Ablation}
%\paragraph{Do we need equivariant covariance?} We investigate the importance of equivariant covariance as described in section~\ref{eq_covariance} for both SO(2) and O(2) groups. In Table 
%~\ref{NN_ablation_table} and Table 
%~\ref{EKF_ablation_table}, the Eq Frame SO(2) + 2 scalars and Eq Frame O(2) + 2 scalars represent the models trained with invariant covariance. It is clearly seen that equivariant covariance provides better results when combined with EKF as EKF improves in performance with a better estimate of the measurement covariance.
%\paragraph{Can a full covariance matrix predicted via person parameterization further improve the performance?} The result of pearson parameterization of the covariance matrix is in inline with our assumption in Section~\ref{eq_covariance}.  We further present visualization of covariance consistency of our Eq Frame O(2) model with diagonal covariance in the equivariant frame in Appendix~\ref{covariance_consistency}.
%Lastly, Appendix~\ref{aug_test_results} contains the results of evaluating separately all the above models on test dataset augmented with rotations and/or reflections.
\textit{\textbf{Covariance Ablation: }}\textbf{Do we need equivariant covariance?} We investigated the importance of equivariant covariance for both $SO(2)$ and $O(2)$ groups, as described in Sec.~\ref{equivariant_frame}(See App.~\ref{covariance_pre} for covariance parameterizations). In Tab.~\ref{NN_ablation_table}, the models +S (rows 6 and 9) are trained with invariant covariance parameterized as $\Sigma=\text{diag}(e^{2u_x}, e^{2u_y}, e^{2u_z})$, that is unaffected by application of $F$. The results show that the equivariant covariance yields better performance, especially when combined with EKF, as it provides a more accurate estimate of the prediction covariance. \textbf{Can a full covariance matrix predicted via Pearson parameterization further improve the performance?} In Tab.~\ref{NN_ablation_table}, +P (rows 7 and 10) are outperformed by our model in most cases. This experimental result indicates that by aligning the principle axis of covariance into the basis of the equivariant frame, we intrinsically force covariance in the equivariant frame to be diagonal, which reduces ambiguity while training. Diagonal covariances improves convergence stability during optimization as stated in~\cite{liu2020tlio}. App.~\ref{covariance_consistency} visualizes the covariance consistency of EqNIO.

\textit{\textbf{Gravity Sensitivity Analysis:}} As ~\citet{wang2023surprisingeffectivenessequivariantmodels} we study the robustness of $SO(3)$ subequivariance to slight gravity axis perturbations in Fig.~\ref{ate_box_grav}. We perturb the gravity axis in the test set by uniformly sampled angles $\alpha\sim\mathcal{U}(-\theta, \theta)$, where $\theta=2^\circ,4^\circ,6^\circ,8^\circ$. As ~\cite{liu2020tlio} we observe that ATE*(m) can be stabilized by training with $\theta=5$ gravity perturbation augmentation. The error of EqNIO trained without this augmentation increases with increasing degree of perturbation.

% \begin{itemize}
% \item covariance ablation on both the groups - pearson parameterization has ambiguity and hence we can see decreased performance - both so(2) and o(2) on TLIO and Aria - NN and EKF - we can also have covariance 3sigma consistency plot here for our full covariance matrix 
% \item in most cases full cov is better than the other two
% \item inline with our assumption that pearson is ambiguous
% \end{itemize}

% \begin{itemize}
%     \item we further do test augmentation on all ablation networks
%     \item refer appendix for random rotations +ref ablation
%     \item this further proves that our equivariance method is superior
%     \item other methods performance decreased while ours is stable and robust
% \end{itemize}
\vspace{-5mm}
\section{Conclusion}
\vspace{-3mm}
We introduce a robust and generalizable neural displacement prior that combats the drift of IMU-only neural inertial odometry by leveraging effective, equivariant canonicalization.
Our canonicalization scheme respects eliminates the underlying yaw ambiguity in gravity-aligned frames which arise from roto-reflections in the plane around gravity. We fully characterize actions from this group on all relevant inputs and outputs of the prior and leverage this insight to design a neural network that produces yaw frames that are $O_g(3)$ equivariant to these actions. As a result, it reduces the data variability seen by existing neural networks, which can be simply integrated into our framework. Compared to existing methods that enforce this equivariance via auxiliary losses or data augmentation, we do this \emph{exactly}, and demonstrate greater generality and versatility through extensive validation on various datasets, and application to two base architectures (TLIO and RONIN). 
We believe this work paves the way for robust, and low-drift odometry running on edge devices.

\bibliography{iclr2025_conference}
\bibliographystyle{iclr2025_conference}

% \appendix
% % \section{Appendix}
\section{Appendix}
\subsection{Preliminary}
\subsubsection{Equivariance}
\label{equi_pre}
In this section, we introduce more preliminaries of group and representation theory which form the mathematical tools for equivariance. 
\paragraph{Group} The group $G$ is a set equipped with an associative binary operation $\cdot$ which maps two arbitrary two elements in $G$ to an element in $G$. It includes an identity element, and every element in the set has an inverse element.

In this paper, we focus on the group $SO(2)$ and $O(2)$. $SO(2)$ is the set of all 2D planar rotations, represented by orthogonal matrices $R_{2\times 2}\in\mathbb{R}^{2\times 2}$ with $\text{det}(R_{2\times 2})=1$. This group operation is matrix multiplication, and each rotation matrix has an inverse, which is its transpose. The identity element is the matrix representing no rotation. 

$O(2)$ consists of all distance-preserving transformations in Euclidean 2D space, including both rotations and reflections. Elements of $O(2)$ are orthogonal matrices $R_{2\times 2}\in\mathbb{R}^{2\times 2}$ with $\text{det}(R_{2\times 2})=\pm 1$, with the group operation being matrix multiplication. Each transformation matrix has an inverse, and the identity element is the matrix representing no transformation.

\paragraph{Group Representation and Irreducible Representation} 
The group representation is a homomorphism from the group $G$ to the general linear map of a vector space $V$ of a field $K$, denoted $GL(V)$. In this work, we use $V=\mathbb{R}^n$ and $K=\mathbb{R}$ 

An irreducible representation (irrep) of a group $G$ is a representation in which the only invariant subspaces under the action of $G$ are the trivial subspace $\{\bm{0}\}$ and the entire space $V$.  In other words, an irreducible representation cannot be broken down into smaller, nontrivial representations,i.e., it cannot be the direct sum of several nontrivial representations. 

For $SO(2)$, we can use $\theta \in (0, 2\pi ]$ to represent $SO(2)$, for any $\theta$, 
the irreducible representation of the frequency $n \in \mathbb{N}$ is: 
\begin{align*}
\rho_n(\theta) = \begin{pmatrix}
\cos n\theta & -\sin n\theta \\
\sin n\theta & \cos n\theta
\end{pmatrix}.
\end{align*}
In this work we use $n=1$. For $O(2)$,  we can use $r \in \{-1,1\}$ to denote reflection and $\theta \in (0, 2\pi]$  to denote rotation. The trivial representation $\rho_0 (r, \theta)=1$. For the nontrivial representation of frequency $n \in \mathbb{N}^{+}$

\[
\begin{aligned}
    &\rho_{n}(r,\theta) = \begin{pmatrix}
    \cos(n\theta) & -\sin(n\theta) \\
    \sin(n\theta) & \cos(n\theta)
    \end{pmatrix} \begin{pmatrix}
    1 & 0 \\
    0 & r
    \end{pmatrix} 
\end{aligned}
\]

There is another one-dimensional irreps for $O(2)$, $\rho(r,\theta) = r$ which corresponds to the trivial representation of rotation.

The introduction to group representations has been covered extensively in previous work on equivariance~\citep{cohen2016group, weiler20183d, xu2024se}. Specifically, for $SO(2)$ and $O(2)$,~\citet{weiler2019general} provide a detailed introduction.

\paragraph{Invariance and Equivariance}
Given a network $\Phi: \mathcal{X} \rightarrow \mathcal{Y}$, if for any $x \in \mathcal{X}$,
\begin{align*}
\Phi (\rho^{\mathcal{X}} x) = \Phi(x),
\end{align*}
implies the group representation $\rho^{\mathcal{Y}}$ of the output space is trivial, \emph{i.e.} identity, and the input does not transform (\emph{i.e.} the input is invariant) under the action of the group. Note that in our paper $\mathcal{X}$ and $\mathcal{Y}$ can be seen as the direct sum (concatenation) of all input and output vectors respectively.
In our paper, the coordinates/ projections of $3D$ vector to the gravity axis $z$-axis are invariant, therefore we call them invariant scalars.

A network $\Phi: \mathcal{X} \rightarrow \mathcal{Y}$ is equivariant if it satisfies the constraint 
\begin{align*}
\Phi (\rho^{\mathcal{X}} x) = \rho^{\mathcal{Y}}\Phi(x).
\end{align*}

In this paper, the input is the sequence of accelerations and angular velocities, and the output are composed of displacement and covariance. For displacement,  the $z$-component is invariant while $xy$-components are acted under the representation of $\rho_1$ defined in the above section. Hence, for displacement, $\rho^{\mathcal{Y}}=\rho_1 \oplus 1$ and for covariance $3D$ covariance, $\rho^{\mathcal{Y}}=(\rho_1 \oplus 1) \otimes (\rho_1 \oplus 1) $

% When a network $\Phi: \mathcal{X} \rightarrow \mathcal{Y}$, we have for any $x \in \mathcal{X}$,
% \begin{align*}
% \Phi (\rho^{\mathcal{X}} x) = \Phi(x),
% \end{align*}
% which means the group representation $\rho^{\mathcal{Y}}$ of the output space is trivial, i.e. identity.
% In our paper, the coordinates/ projections of $3D$ vector to the gravity axis $z$-axis are invariant, therefore we call them invariant scalars.

%  The network $\Phi: \mathcal{X} \rightarrow \mathcal{Y}$ is equivariant means it satisfying that 
% \begin{align*}
% \Phi (\rho^{\mathcal{X}} x) = \rho^{\mathcal{Y}}\Phi(x).
% \end{align*}

% In this paper, when the output is displacement since the $z-$ coordinate is invariant while $xy-$ displacement is under representation of $\rho_1$ defined in the above section, we have $\rho^{\mathcal{Y}}=\rho_1 \oplus 1$; for $3D$ covariance, we have $\rho^{\mathcal{Y}}=(\rho_1 \oplus 1) \otimes (\rho_1 \oplus 1) $

\paragraph{Subequivariance}
As mentioned in prior works~\citep{chen2023subequivariant,han2022learning}, the existence of gravity breaks the symmetry in the vertical direction, reducing O(3) to its subgroup O(2). We formally characterize this phenomenon of equivariance relaxation as subequivariance. We have mathematically defined the subequivariance in Section~\ref{problem_setup} of the paper. In simpler terms, the gravity axis is decoupled and treated as an invariant scalar while the other two axes are handled as a separate 2D vector. Upon rotation, the invariant scalar remains constant while the other two axes are transformed under rotation. So we are limited now to SO(2) rotations and roto-reflections. In the general case of equivariance, the 3D vector would be considered three-dimensional and an SO(3) rotation would act on it. The transformation would be along all three axes. 

\subsubsection{Inertial Odometry}
\label{inertial_odomerty_pre}
In this section, we introduce more preliminaries on the terms used in inertial odometry.
\paragraph{Inertial Measurement Unit} Inertial Measurement Unit (IMU) is an electronic device that measures and reports linear acceleration, angular velocity, orientation, and other gravitational forces. An IMU typically consists of a 3-axis accelerometer, a 3-axis gyroscope, and depending on the heading requirement a 3-axis magnetometer. 

An accelerometer measures instantaneous linear acceleration ($a_i$). It can be thought of as a mass on a spring, however in micro-electro-mechanical systems (MEMS) it is beams that flex instead of spring.

A gyroscope measures instantaneous angular velocity ($\omega_i$). It measures the angular velocity of its frame, not any external forces. Traditionally, this can be measured by the fictitious forces that act on a moving object brought about by the Coriolis effect, when the frame of reference is rotating. In MEMs, however, we use high-frequency oscillations of a mass to capture angular velocity readings by the capacitance sense cones that pick up the torque that gets generated.

\paragraph{World Frame}
A world frame, also known as a cartesian coordinate frame, is a fixed frame with a known location and does not change over time. In this paper, we denote the fixed frame with $z$-axis perfectly aligned with the gravity vector as the world frame, denoted as $w$.

\paragraph{Local-gravity-aligned Frame}
A local-gravity-aligned frame has one of its axes aligned with the gravity vector at all times but it is not fixed to a known location. 

\paragraph{Body Frame}
A body frame comprises the origin and orientation of the object described by the navigation solution. In this paper, the body frame is the IMU's frame. This is denoted as $b$ for the IMU data.

\paragraph{Gravity-compensation}
Gravity compensation refers to the removal of the gravity vector from the accelerometer reading.

\paragraph{Gravity-alignment}
Gravity-alignment of IMU data refers to expressing the data in the gravity-aligned frame. This is done by aligning the $z$-axis of the IMU inertial frame with the gravity vector pointing downwards and is usually achieved by fixing the roll and pitch (rotations around the $x$ and $y$ axes) or by applying a transformation estimated by the relative orientation between the gravity vector and a fixed $z$-axis pointing downwards. This is usually achieved with a simple rotation.

\subsubsection{Uncertainty Quantification in Inertial Odometry}
\label{covariance_pre}
In this section, we provide more context on uncertainty quantification in odometry and detail the different parameterizations used for regressing the covariance matrix in the paper.
\paragraph{Homoscedastic Uncertainty}
Homoscedatic uncertainty refers to uncertainty that does not vary for different samples, i.e., it is constant.

\paragraph{Heteroscedastic Uncertainty}
Heteroscedastic uncertainty is uncertainty that is dependent on the sample, i.e., it varies from sample to sample.

\paragraph{Epistemic Uncertainty}
Epistemic uncertainty is uncertainty in model parameters. This can be reduced by training the model for longer and/or increasing the training dataset to include more diverse samples.

\paragraph{Aleatoric Uncertainty}
Aleatoric uncertainty is the inherent noise of the samples. This cannot be reduced by tuning the network or increasing the diversity of the data.

\paragraph{Why do we need to estimate uncertainty in inertial odometry?}
In inertial odometry when we use a probabilistic filter-based approach like a Kalman Filter, the filter estimates the probability distribution over the pose recursively. While integrating the neural network prediction, the filter fuses the prediction with other sensor measurements, like raw IMU data in TLIO~\citep{liu2020tlio}, by weighing it based on the accuracy or reliability of the measurements. For neural networks, this reliability is obtained by estimating the uncertainty. If we use a fixed uncertainty (homoscedastic) it is seen to cause catastrophic failures of perception systems. The uncertainty estimated in TLIO captures the extend to which input measurements encode the motion model prior.

\paragraph{What is the uncertainty we are estimating in inertial odometry?}We are regressing aleatoric uncertainty using the neural network and training the model till the epistemic uncertainty is very small as compared to aleatoric uncertainty.

\paragraph{How is the uncertainty estimated in this paper?} We regress aleatoric uncertainty as a covariance matrix jointly while regressing 3D displacement following the architecture of TLIO~\citep{liu2020tlio}. Since there is no ground truth for the covariance, we use the negative log-likelihood loss of the prediction using the regressed Gaussian distribution. As this loss captures the Mahalanobis distance, the network gets jointly trained to tune the covariance prediction. We do not estimate epistemic uncertainty separately in this paper, but as mentioned in~\citet{russell2021multivariate} we train the network until the epistemic uncertainty is small as compared to aleatoric uncertainty.

\paragraph{Diagonal covariance matrix}
TLIO~\citep{liu2020tlio} regresses only the three diagonal elements of the covariance matrix as log $\sqrt{\Sigma_{xx}}$, log $\sqrt{\Sigma_{yy}}$ and log $\sqrt{\Sigma_{zz}}$ and the off-diagonal elements are zero. This formulation assumes the axes are decoupled and constrains the uncertainty ellipsoid to be along the local gravity-aligned frame. 

\paragraph{Full covariance matrix using Pearson correlation}
\citet{russell2021multivariate} define a parameterization to regress the full covariance matrix. They regress six values of which three are the diagonal elements log $\sqrt{\Sigma_{xx}}$, log $\sqrt{\Sigma_{yy}}$ and log $\sqrt{\Sigma_{zz}}$ and the remaining three are Pearson correlation coefficients $\rho_{xy}$, $\rho_{yz}$, and $\rho_{xz}$. The diagonal elements are obtained by exponential activation while the off-diagonal elements are computed as follows 
\[
\Sigma_{ij} = \rho_{ij}\sqrt{\Sigma_{ii}\Sigma_{jj}}
\]
where $\rho_{ij}$ passes through tanh activation.

\paragraph{Diagonal covariance matrix in canonical frame}
In our approach, we regress the three diagonal elements as log $\sqrt{\Sigma_{xx}}$, log $\sqrt{\Sigma_{yy}}$ and log $\sqrt{\Sigma_{zz}}$ in the invariant canonical frame. Since the $z$-axis is decoupled from the $xy$-axis, only $\Sigma_{xx}$ and $\Sigma_{yy}$ are mapped back using the equivariant frame to obtain a full 2D covariance matrix from the diagonal entries. The resulting matrix is as follows
\[
\begin{bmatrix}
    \Sigma_{xx} & \Sigma_{xy} & 0 \\
    \Sigma_{xy} & \Sigma_{yy} & 0 \\
    0 & 0 & \Sigma_{zz}
    \end{bmatrix}
\]
\subsection{Dataset Details}
\label{dataset_details}
In this section, we provide a detailed description of the 4 datasets used in this work - TLIO and Aria for TLIO architecture, and RONIN, RIDI and OxIOD for RONIN architecture.
\paragraph{TLIO Dataset- } The TLIO Dataset~\citep{liu2020tlio} is a headset dataset that consists of IMU raw data at 1kHz and ground truth obtained from MSCKF at 200 Hz for 400 sequences totaling 60 hours. The ground truth consists of position, orientation, velocity, IMU biases and noises in $\mathbb{R}^{3}$. The dataset was collected using a custom rig where an IMU (Bosch BMI055) is mounted on a headset rigidly attached to the cameras. This dataset captures a variety of activities including walking, organizing the kitchen, going up and down stairs, on multiple different physical devices and more than 5 people for a wide range of individual motion patterns, and IMU systematic errors. We use their data splits for training (80\%), validation (10\%), and testing(10\%).

\paragraph{Aria Everyday Dataset- } Aria Everyday Dataset~\citep{lv2024aria} is an open-sourced egocentric dataset that is collected using Project Aria Glasses. This dataset consists of 143 recordings accumulating to 7.3 hrs capturing diversity in wearers and everyday activities like reading, morning exercise, and relaxing. There are two IMUs on the left and right side of the headset of frequencies 800 and 1kHz respectively. They have two sources of ground truth- open and closed loop trajectory at 1kHz. Open loop trajectory is strictly causal while closed loop jointly processes multiple recordings to place them in a common coordinate system. The ground truth contains position and orientation in $\mathbb{R}^{3}$. We use it as a test dataset. The raw right IMU data is used to compare closed-loop trajectory with EKF results. The data was downsampled to 200Hz and preprocessed using the closed-loop trajectory to test the Neural Network trained on TLIO.

\paragraph{RONIN Dataset- } RONIN Dataset~\citep{herath2020ronin} consists of pedestrian data with IMU frequency and ground truth at 200Hz. RONIN data features diverse sensor placements, like the device placed in a bag, held in hand, and placed deep inside the pocket, and multiple Android devices from three vendors Asus Zenfone AR, Samsung Galaxy S9 and Google Pixel 2 XL. Hence, this dataset has different IMUs depending on the vendor. We use RONIN data splits to train and test their model with and without our framework.

\paragraph{RIDI Dataset- } RIDI Dataset~\citep{Yan_2018_ECCV} is another pedestrian dataset with IMU frequency and ground truth at 200 Hz. This dataset features specific human motion patterns like walking forward/backward, walking sidewards, and acceleration/deceleration. They also record data with four different sensor placements. We report test results of RONIN models on both RIDI test and cross-subject datasets. RIDI results are presented after post-processing the predicted trajectory with the Umeyama algorithm~\citep{umeyama1991} for fair comparison against other methods. 

\paragraph{OxIOD Dataset- } OxIOD Dataset~\citep{OXOID} stands for  Oxford Inertial Odometry Dataset consists of various device placements/attachments, motion modes, devices, and users capturing everyday usage of mobile devices. The dataset contains 158 sequences totaling 42.5 km and 14.72 hours captured in a motion capture system. We use their unseen multi-attachments test dataset for evaluating our framework applied to RONIN architecture.

\subsection{Equivariant Network Implementation Details}
\label{network_details}
In this section, we describe in detail the equivariant network implementation and how it is combined with TLIO and RONIN. The input to the framework is IMU samples from the accelerometer and gyroscope for a window of 1s with IMU frequency 200Hz resulting in $n$ = 200 samples. All IMU samples within a window are gravity-aligned with the first sample at the beginning of the window, previously referred to as the clone state. The network design, as seen in Figure~\ref{pipeline} b, differs in architecture for $SO(2)$ and $O(2)$ and hence described separately below.
\paragraph{$SO(2)$-} We decouple the $z$-axis from the other two axes and treat linear acceleration and angular velocity along the $z$-axis as scalars (2). We also take the norm of the 2D accelerometer and gyroscope measurements (2), their inner product (1) resulting in invariant scalars $\mathbb{R}^{ n \times 
 5}$. The $x$ and $y$ components of IMU measurements are passed as vector inputs $\mathbb{R}^{n \times 2 \times 2 }$. The vectors and scalars are then separately passed to the linear layer described in Section~\ref{basic_layers}. The equivariant network predicting the equivariant frame consists of 1 linear layer, 1 nonlinearity, 1 convolutional block with convolution applied over time, non-linearity, and layer norm. The hidden dimension is 128 and the convolutional kernel is 16 x 1. Finally, the fully connected block of hidden dimension 128 and consisting of linear, nonlinearity, layer norm, and output linear layer follows a pooling over the time dimension. The output of the final linear layer is 2 vectors representing the two bases of the equivariant frame. The input vectors of dimension $\mathbb{R}^{ n \times 2 \times 2 }$ are projected into the invariant space via the equivariant frame resulting in invariant features in $\mathbb{R}^{ n \times 4 }$. These features are combined with the input scalars and passed as input ($\mathbb{R}^{ n \times 6 }$) to TLIO or RONIN base architecture. The output of TLIO is invariant 3D displacement and diagonal covariance along the principal axis. The output of RONIN is 2D velocity. The $x$ and $y$ components are back-projected using the equivariant frame to obtain displacement vector \textit{d} in $\mathbb{R}^{2}$ and the covariance in the original frame. The covariance is parameterized and processed as mentioned in Section~\ref{equivariant_frame}.

\paragraph{$O(2)$- } The preprocessing is as described in Section~\ref{equivariant_frame} where $\omega$ is decomposed to two vectors $v_1$ and $v_2$ that each have magnitude $\sqrt{\Vert \omega \Vert}$. The preprocessed input therefore consists of 3 vectors 
$a$, $v_1$ and $v_2$. This is then passed to the equivariant network by decoupling the $z$-axis resulting in vector input $\mathbb{R}^{ n \times 3 \times 2 }$ which represents 3 vectors in 2D. The scalars passed to the linear layer described in Section~\ref{basic_layers} consist of the accelerometer $z$-axis measurement (1), the $z$ component of the two vectors $v_1$ and $v_2$ (2), the norm of the vectors (3) and the inner product of the vectors(3) resulting in $\mathbb{R}^{ n \times 9 }$. The network architecture is the same as $SO(2)$ with hidden dimension 64 and 2 convolutional blocks in order to make it comparable in the number of parameters to $SO(2)$ architecture. The invariant features obtained by projecting the three vectors using the equivariant frame are processed as mentioned in Section~\ref{equivariant_frame} to obtain 2 vectors in 3D that are fed as input to TLIO and RONIN. The postprocessing is the same as $SO(2)$.

The framework is implemented in Pytorch and all hyperparameters of the base architectures are used to train TLIO and RONIN respectively. The $SO(2)$ architecture has 8,884,870 while $O(2)$ has 6,020,230 number of parameters and the base TLIO architecture has 5,424,646. The baseline TLIO and our methods applied to TLIO were trained on NVIDIA a40 GPU occupying 7-8 GB memory per epoch. The training took 5 mins per epoch over the whole training dataset. We train for 10 epochs with MSE Loss and the remaining 40 epochs with MLE Loss similar to TLIO~\citep{liu2020tlio}. RONIN was trained on NVIDIA 2080ti for 38 epochs taking 2 mins per epoch. The loss function used was MSE as mentioned in~\citet{herath2020ronin}. The EKF described in TLIO was run on NVIDIA 2080ti with the same initialization and scaling of predicted measurement covariance as in TLIO~\citep{liu2020tlio}.

We compare the resource requirements of the SO(2), and O(2) variant of our method coupled with TLIO, with base TLIO without an equivariant frame. We report the floating point operations (FLOPs), the inference time (in milliseconds), and Maximum GPU memory (in GB) during inference, on an NVIDIA 2080 Ti GPU for the neural network averaged over multiple runs to get accurate results. While base TLIO uses 35.5 MFLOPs, 3.5 ms, and 0.383 GB per inference, our SO(2) equivariant method instead uses 531.9 MFLOPs, 4.3 ms, and 0.383 GB per inference. Finally, our O(2) equivariant method uses 638.5 MFLOPs, 4.6 ms, and 0.385 GB per inference. We further evaluate the Maximum GPU memory for the equivariant networks separately and report 0.255 GB per inference for SO(2) equivariant frame prediction and 0.257 GB per inference for O(2) equivariant frame prediction. The Maximum GPU memory is unaffected because the equivariant frame computation utilizes less memory than TLIO.

Finally, we also evaluate our method with a downstream EKF on an NVIDIA 2080 Ti GPU. The EKF incorporates raw IMU measurements for propagation, and displacement measurements from the neural network as measurement updates. For every 20 imu samples, we send the last 200 IMU measurements to the neural network to provide this measurement update. The original TLIO requires 0.492 seconds and 1.113 GB of memory. For the SO(2) variant of our method, we require 0.554 seconds and 1.109 GB of memory to process 1 second of real-world data. For the O(2) variant, we use 0.554 seconds and 1.115 GB of memory, showing that our method is faster than real-time. The increase in memory for the O(2) variant is due to the additional preprocessing step.

With comparable computing resources, our equivariant model outperforms TLIO since we leverage symmetry, which is an intrinsic property in inertial odometry.

\subsection{EKF Details}
\label{App:EKF}
The EKF continuously estimates the orientation, linear velocity, position, acceleration, and gyroscope biases. In TLIO architecture, the EKF propagates the IMU samples at a higher frequency of 1 kHz while the neural network takes clone states at 200Hz to predict the displacement over 1s time window and its associated uncertainty. Throughout the paper, we denote the clone state indices with i and the EKF propagation indices with k. We further outline the EKF state definition, propagation model, state augmentation, measurement, and update model below for completeness of the manuscript. It must be noted that we follow TLIO~\citep{liu2020tlio} and do not make any modifications to this part of the architecture.
\subsubsection{EKF State Definition}
The state of the EKF is defined as $X = (\xi_1, ...., \xi_n, s)$ where $\xi_i, i=1,...,n$ represents the $n$ clone states whose corresponding IMU measurements are passed to the neural network as input and $s$ has the current propagation state. 
\[
\xi_i = (\prescript{w}{b}{R_i}, \prescript{w}{}{p_i}) \quad s = (\prescript{w}{b}{R}, \prescript{w}{}{v}, \prescript{w}{}{p}, b_g, b_a)
\]
where $\prescript{w}{b}{R_i}$ represents the orientation estimate of EKF from IMU body frame to gravity-aligned world frame, $p$ and $v$ represent the position and linear velocity estimates, and $b_g$ and $b_a$ denote the bias estimates of the gyroscope and accelerometer respectively. The error state of the EKF is propagated as a linearized error and so the error state is defined as 
\[
\tilde{\xi_i} = (\tilde{\theta_i}, \delta\tilde{p_i}) \quad \tilde{s} = (\tilde{\theta}, \delta\tilde{v}, \delta\tilde{p}, \delta\tilde{b_g}, \delta\tilde{b_a})
\]
where tilde indicates errors in every state. The errors for all states, except orientation, are approximated to simple subtraction even though the $SE(3)$ parameterization we use in the EKF process model is $T(3) \times SO(3)$. For orientation we use the logarithm map of rotation to find the error as $\tilde{\theta} = log_{SO(3)} (R\hat{R}^{-1}) \in \mathfrak{so}(3)$. The EKF and error state is of dimension (6n + 15) and the corresponding state covariance matrix P has dimension (6n + 15, 6n + 15).
\subsubsection{Process Model}
IMU's measure sequences of data $\{(a_k, \omega_k)\}_{i=k}^m$, each expressed in the local IMU frame at time $t_k$.
These are related to the true IMU acceleration $\bar{a}_k$ and angular rates $\bar{\omega}_k$ via 
\begin{align}
    \quad\omega_k = \bar{\omega}_k + b_k^g + \eta^g_k\quad
   \quad a_k = \bar{a}_k - \prescript{w}{b}{R_k}^T \vec{g} + b_k^a + \eta^a_k\quad
\end{align}
where $\vec{g}$ is gravity vector pointing downward in world frame, and $\eta^g_k$ and $\eta^a_k$ are IMU noises respectively.
The EKF propagation uses raw IMU samples in the local IMU frame $b$, following strap-down inertial kinematics equations:
\begin{gather*}
\small
    \mathbf{\prescript{w}{b}{\hat{R}_{k+1}}} = \mathbf{\prescript{w}{b}{\hat{R}_k}} \exp_{SO(3)}(\mathbf{(\omega_{k} - \hat{b}_{gk})}\Delta t)\\
    \mathbf{\prescript{w}{}{\hat{v}_{k+1}}} = \mathbf{\prescript{w}{}{\hat{v}_k}} + \mathbf{\prescript{w}{}{g}}\Delta t + \mathbf{\prescript{w}{b}{\hat{R}_k}(a_k - \hat{b}_{ak})}\Delta t\\
    \mathbf{\prescript{w}{}{\hat{p}_{k+1}}} = \mathbf{\prescript{w}{}{\hat{p}_k}} + \mathbf{\prescript{w}{}{\hat{v}_k}}\Delta t + \frac{1}{2} \Delta t^2 (\mathbf{\prescript{w}{}{g} + \mathbf{\prescript{w}{b}{\hat{R}_k}}(a_k - \hat{b}_ak)})\\
    \mathbf{\hat{b}_{g(k+1)} = \hat{b}_{gk} + \eta_{gdk}}\\
    \mathbf{\hat{b}_{a(k+1)} = \hat{a}_{gk} + \eta_{adk}}
\end{gather*}
where at timestep \textit{k}, $\Delta t $ is the time interval, $\prescript{w}{}{g}$ is the constant gravity vector, $\eta_{gdk}$ and $\eta_{adk}$ are the IMU noises that are assumed to be normally distributed, and $\exp_{SO(3)}$ is the $SO(3)$ exponential map.

The linearized error propagation is written as:
\[
\tilde{s_{k+1}} = A_{k(15,15)}^s \tilde{s_k} + B_{k(15,12)}^s n_k
\]
where $n_k = [n_{\omega k},n_{ak},\eta_{gdk}, \eta_{adk}]$ and the subscript brackets (.,.) indicate matrix dimensions. The corresponding linearized propagation of the state covariance P is as follows:
\[
P_{k+1} = A_k P_k A_k^T + B_k W B_k^T
\]
\[
A_k = \begin{bmatrix}
I_{6n} & 0 \\
0 & A_k^s
\end{bmatrix} \quad
B_k = \begin{bmatrix}
0 \\
B_k^s
\end{bmatrix}
\]
where $I$ stands for identity matrix amd $W_{(12, 12)}$ is the covariance matrix of sensor noise and bias random walk.
\subsubsection{State Augmentation}
As the neural network inputs are used to correct the EKF estimates at a frequency of 20Hz while the propagation of the EKF is at 1kHz as mentioned in TLIO, at the measurement update frequency a new state is augmented with a copy operation incrementing the dimension as seen below:
\[
P_{k+1} = \bar{A_k} P_k \bar{A_k^T} + \bar{B_k} W \bar{B_k^T}
\]
\[
\bar{A_k} = \begin{bmatrix}
I_{6n} & 0 \\
0 & A_k^\xi\\
0 & A_k^s
\end{bmatrix} \quad
\bar{B_k} = \begin{bmatrix}
0 \\
B_k^\xi\\
B_k^s
\end{bmatrix}
\]
where $A_k^\xi$ and $B_k^\xi$ are the partial propagation matrices for rotation and position only with dimension (6, 15) and (6, 12) respectively.
\subsubsection{Measurement Model}
% \subsection*{Measurement Model}
The measurement model in the EKF uses the displacement estimates provided by the neural network, aligning them in a local gravity-aligned frame to ensure the measurements are decoupled from global yaw information:
\[
\hat{h}(\mathbf{X}) = \mathbf{R}_\gamma^T (\mathbf{p}_j - \mathbf{p}_i) = \hat{d}_{ij} + \eta_{ij}
\]
where $\mathbf{R}_\gamma$ is the yaw rotation matrix, $\mathbf{p}_i$ and $\mathbf{p}_j$ are positions of the first and last clone state for the 1s displacement prediction window, $\hat{d}_{ij}$ indicates the predicted displacement which is the output of the neural network and $\eta_{ij}$ represents the measurement noise modeled by the network's uncertainty output as $\eta_{ij} = \mathcal{N}(0,\hat{\Sigma}_{ij})$. The orientation estimate of the EKF at clone state i is decomposed using extrinsic "XYZ" Euler angle convention as $R_i = R_{\gamma} R_{\beta} R_{\alpha}$. 

\subsubsection{Update Model}
% \subsection*{Update Step}
The Kalman gain is computed based on the measurement and covariance matrices, and the state and covariance are updated accordingly. The key update equations involve the computation of the Kalman gain ($\mathbf{K}$), updating the state ($\mathbf{X}$), and updating the covariance matrix ($\mathbf{P}$):
\[
\mathbf{K} = \mathbf{P} \mathbf{H}^T (\mathbf{H} \mathbf{P} \mathbf{H}^T + \hat{\Sigma}_{ij})^{-1}
\]
\[
\mathbf{X} \longleftarrow \mathbf{X} \oplus \mathbf{K}(h(\mathbf{X}) - \hat{d}_{ij})
\]
\[
\mathbf{P} \longleftarrow (\mathbf{I} - \mathbf{K} \mathbf{H}) \mathbf{P}(\mathbf{I} - \mathbf{K} \mathbf{H})^T +\mathbf{K}\hat{\Sigma}_{ij}\mathbf{K}^T
\]
where $\oplus$ denotes addition operation except for rotation where the update operation writes $R \longleftarrow \exp(\tilde{\theta})R$ and the linearized measurement matrix $\mathbf{H}_{(3, 6n+15)}$ which has zeros other than
\[
\mathbf{H}_{\tilde{\theta_i}} = \frac{\partial h(X)}{\partial \tilde{\theta_i}} = \hat{R_\gamma^T} {\lfloor \prescript{w}{}{\hat{p}_j} - \prescript{w}{}{\hat{p}_i} \rfloor}_\times \mathbf{H}_z
\]
\[
\mathbf{H}_{\delta\tilde{p_i}} = \frac{\partial h(X)}{\partial \delta\tilde{p_i}} = - \hat{R_\gamma^T}
\]
\[
\mathbf{H}_{\delta\tilde{p_j}} = \frac{\partial h(X)}{\partial \delta\tilde{p_j}} = \hat{R_\gamma^T}
\]
where \[
\mathbf{H}_z = \begin{bmatrix}
0 & 0 & 0\\
0 & 0 & 0\\
\cos{\gamma}\tan{\beta} & \sin{\gamma}\tan{\beta}&1
\end{bmatrix}
\] and ${\lfloor x \rfloor}_\times$ is a skew-symmetric matrix built from a vector $x$.
\subsection{Evaluation Metrics Definition}
We follow most metrics in TLIO~\citep{liu2020tlio} and RONIN~\citep{herath2020ronin}, besides $MSE$ loss we reported in the paper. Here we provide the mathematical details of these metrics.
\label{eval_metrics}
\begin{itemize}
    \item MSE (m$^{2}$): Translation error per sample between the predicted and ground truth displacement averaged over the trajectory. It is computed as $\frac{1}{n} \sum^n_i \|\prescript{w}{}{p}_i - \prescript{w}{}{\hat{p}}_i\|$. However, it should be noted that MSE mentioned in TLIO~\citep{liu2020tlio} is the same as MSE Loss calculated as the squared error averaged separately for each axis $\frac{1}{n} \sum^n_i \|\prescript{w}{}{p}_{i,r} - \prescript{w}{}{\hat{p}}_{i,r}\|$ where \textit{r} is an axis.
    \item ATE (m): Translation Error assesses the discrepancy between predicted and ground truth (GT) positions across the entire trajectory. It is computed as $\sqrt{\frac{1}{n} \sum^n_i \|\prescript{w}{}{p}_i - \prescript{w}{}{\hat{p}}_i\|}$
    \item RTE (m): Relative Translation Error measures the local differences between predicted and GT positions over a specified time window of duration $\delta t$ (1 minute). $\sqrt{\frac{1}{n} \sum^n_i \|\prescript{w}{}{p}_{i+\delta t} - \prescript{w}{}{p}_i -(\prescript{w}{}{\hat{p}}_{i+\delta t}-\prescript{w}{}{\hat{p}}_i)\|}$. 
    \item AYE Absolute Yaw Error is calculated as $\sqrt{\frac{1}{n} \sum^n_i \|\gamma_i - \hat{\gamma}_i\|}$.
% In the TLIO experiment, for the Metric-NN, we evaluate the results of the Neural Network, while the metric without the ''-NN" suffix pertains to the results obtained from the Extended Kalman Filter (EKF).In the RONIN experiment, we evaluate all the results of the Neural Network.
\end{itemize}
\subsection{Visualization of TLIO results}
\label{visualization_tlio_main}
\begin{figure}[ht]
\begin{center}
\centerline{\includegraphics[width=\columnwidth]{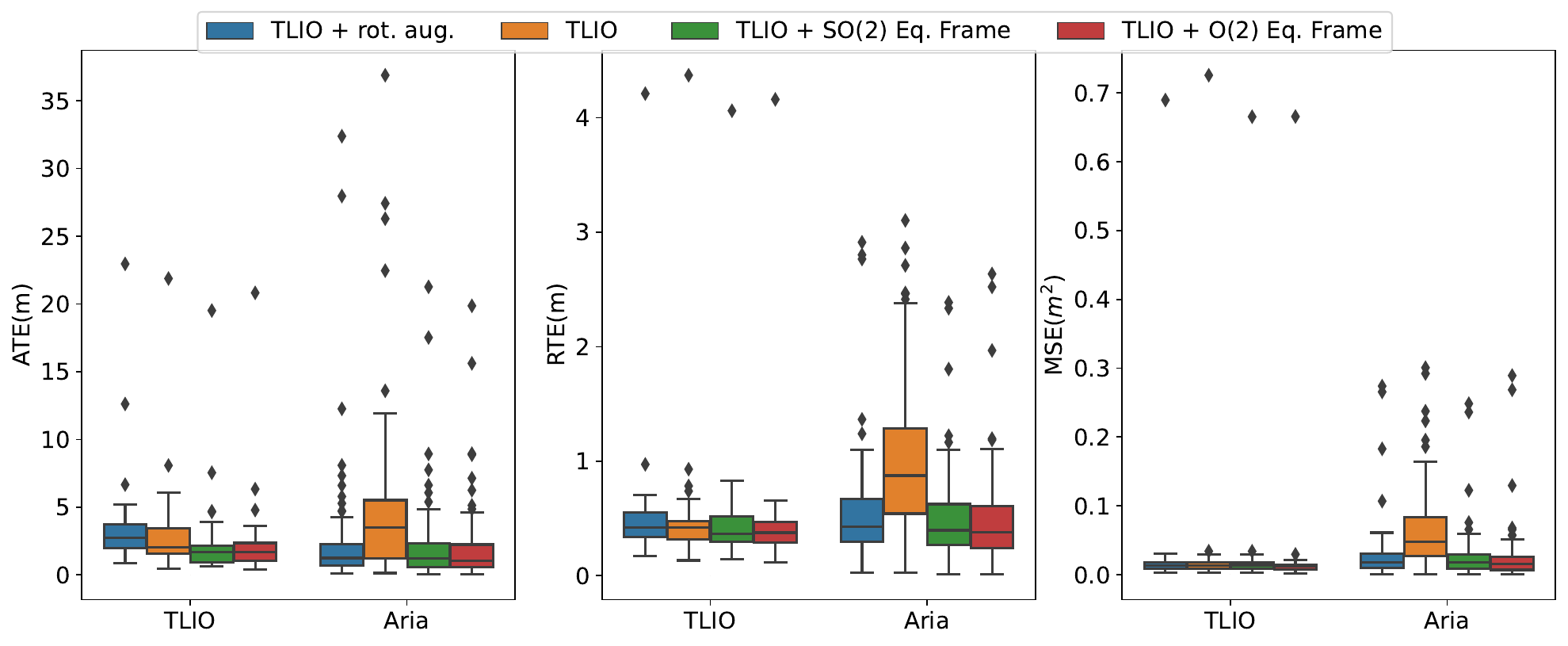}}
\caption{The superior performance of our framework applied to TLIO architecture when compared to baseline TLIO trained with and without augmentations on TLIO and Aria Datasets visualized with a box plot. Blue, Orange, Green and Red indicate +rot. aug., TLIO, +$SO(2)$ Eq. Frame and +$O(2)$ Eq. Frame. ATE, RTE and MSE indicate ATE*, RTE* and MSE* corresponding to only the NN results.}
\label{boxplot_nn_tlio}
\end{center}
\end{figure}
\begin{figure}[ht]
\begin{center}
\centerline{\includegraphics[width=\columnwidth]{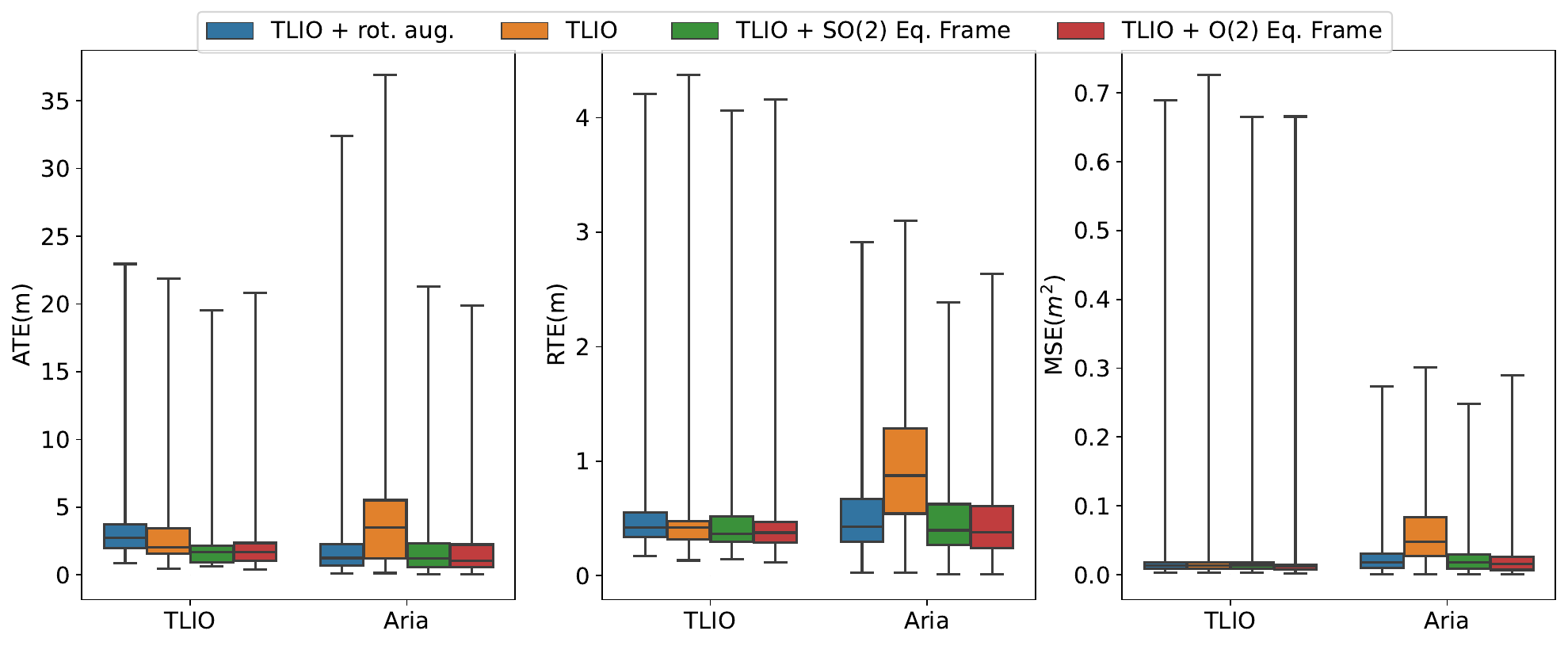}}
\caption{The superior performance of our framework applied to TLIO architecture when compared to baseline TLIO trained with and without augmentations on TLIO and Aria Datasets visualized with a box plot. Blue, Orange, Green and Red indicate +rot. aug., TLIO, +$SO(2)$ Eq. Frame and +$O(2)$ Eq. Frame. ATE, RTE and MSE indicate ATE*, RTE* and MSE* corresponding to only the NN results. The whisker is extended to 1.5 * IQR (inter-quartile range).}
\label{boxplot_nn_tlio_w_outlier}
\end{center}
\end{figure}
\begin{figure}[ht]
\begin{center}
\centerline{\includegraphics[width=\columnwidth]{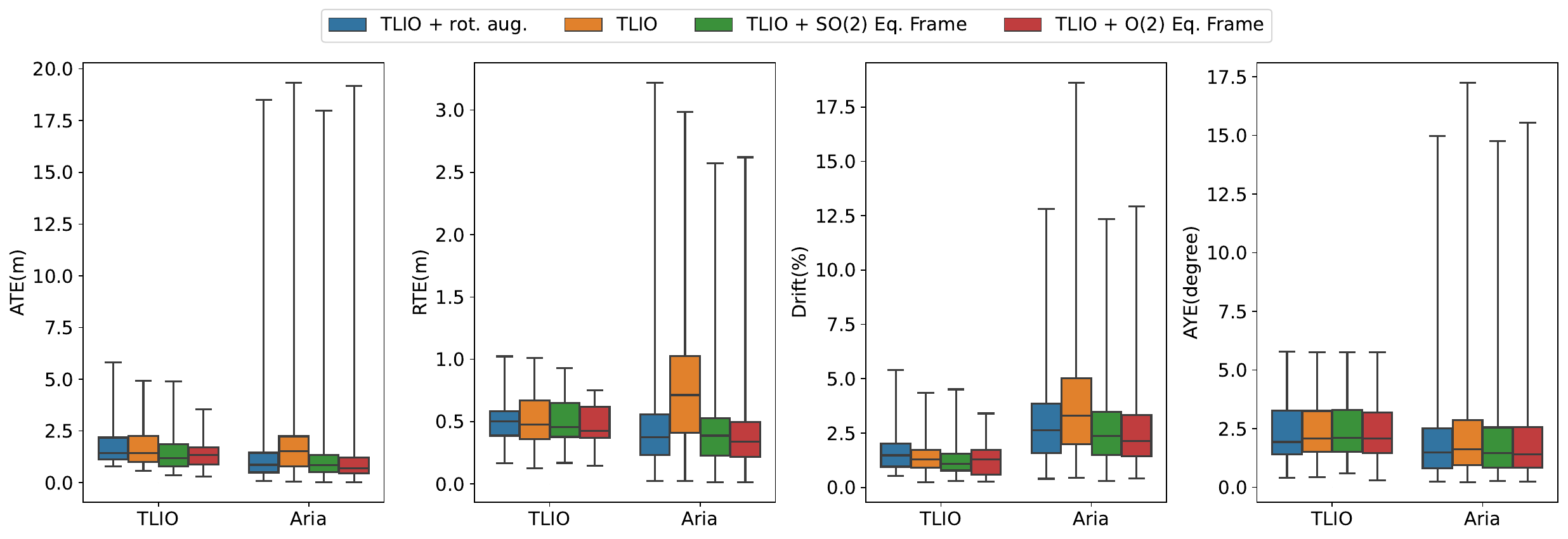}}
\caption{The superior performance of our framework applied to TLIO architecture when compared to baseline TLIO trained with and without augmentations on TLIO and Aria Datasets visualized with a box plot. Blue, Orange, Green and Red indicate +rot. aug., TLIO, +$SO(2)$ Eq. Frame and +$O(2)$ Eq. Frame. The whisker is extended to 1.5 * IQR (inter-quartile range).}
\label{boxplot_ekf_tlio_w_outlier}
\end{center}
\end{figure}
\begin{figure}[ht]
\begin{center}
\centerline{\includegraphics[width=\columnwidth]{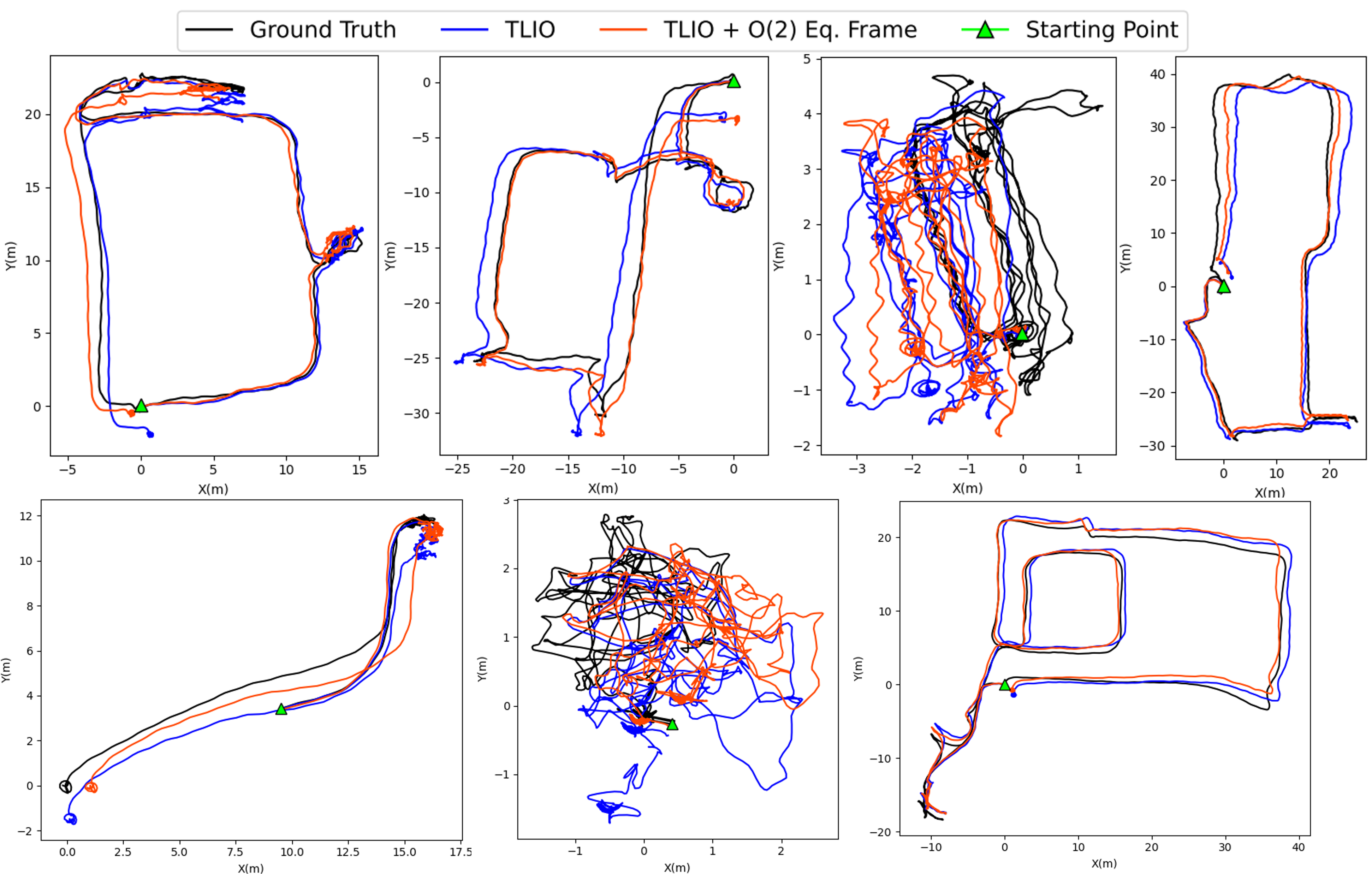}}
\caption{More Visualizations of final estimated trajectories on TLIO Dataset by baseline TLIO (Blue), our best method applied to TLIO (+$O(2)$ Eq. Frame)(Red), and the Ground-Truth trajectory (Black).}
\label{traj_tlio_traj2}
\end{center}
\end{figure}

Figure~\ref{boxplot_nn_tlio} and Figure~\ref{boxplot_nn_tlio_w_outlier} show only the neural network results compared to ground truth displacements. The ATE and RTE is calculated on the cumulative trajectory obtained form the predicted displacements. Figure~\ref{boxplot_nn_tlio_w_outlier} is with whisker extended to include the outlier which are commonly calculated as 1.5 * IQR (inter-quartile range). Figure~\ref{boxplot_ekf_tlio_w_outlier} shows the results of EKF without excluding the outliers. We provide more trajectory visualizations of TLIO test data in Figure~\ref{traj_tlio_traj2}.

\subsection{Augmented TLIO Test Dataset Results and Analysis}
\label{aug_test_results}

We also perform an ablation study on test data augmentation for our model. For neural network results, we apply four random yaw rotations per trajectory and random rotations plus reflection per trajectory. The results are detailed in Table~\ref{aug_net}. Except for our equivariant model, all other methods show decreased performance compared to their results on non-augmented test data, whereas our model maintains consistent performance and outperforms the other methods.

For the Extended Kalman Filter (EKF) results, we augment the test data using random $SO(3)$ rotations. Notably, we do not include reflections due to the structural constraints of the Kalman filter. As shown in Table~\ref{aug_ekf},  despite the +Non Eq. Frame model outperforming ours in non-augmented tests on ATE metrics. Our model exceeds +Non Eq. Frame on the augmented dataset. Our approach not only sets a new benchmark but also maintains consistent performance across random rotations.

\begin{table}[ht]
\newcommand{\first}{\cellcolor{red!40}}
\centering
\begin{tabular}{lcccccc}
\toprule
& \multicolumn{3}{c}{Rotations} & \multicolumn{3}{c}{Rotations + Reflections} \\
\cmidrule(lr){2-4} \cmidrule(lr){5-7}
Model & MSE* & ATE* & RTE* & MSE* & ATE* & RTE*  \\
&($10^{-2}m^2$)&($m$)&($m$)&($10^{-2}m^2$)&($m$)&($m$)\\
\midrule
TLIO & 0.2828 & 27.7797 & 3.1390 & 0.2989 & 23.4839 & 3.1313 \\
+ rot. aug. & 0.0327 & 3.3180 & 0.5417  & 0.0347 & 2.9110 & 0.5654  \\
+ rot. aug. + more layers & 0.0306 & 3.0264 & 0.5300 & 0.0332 & 2.3028 & 0.5592  \\
+ rot. aug. + Non Eq. Frame & 0.0302 & 2.6379 & 0.5025 & 0.0331 & 2.3212 & 0.5446 \\
+ rot. aug. + PCA Frame & 0.2286 & 21.3795 & 2.5288 & 0.2467 & 10.1660 & 2.2283  \\
\hline
+ \textbf{$\mathbf{SO(2)}$ Eq. Frame }  & 0.0319 & \first{2.3218} & 0.4957 & 0.0339 & 1.8664 & 0.5178\\
+ \textbf{$\mathbf{O(2)}$ Eq. Frame } & \first{0.0298} & 2.3305 & \first{0.4719} & \first{0.0298} &\first{1.6418} & \first{0.4361} \\
\bottomrule
\end{tabular}
\caption{Ablation Study For Neural Network with Random Rotation and Reflection Transformation (4 per trajectory) on TLIO test dataset. A lower error indicates a better model. The lowest values are annotated with Red. Our proposed methods are in bold.}
\label{aug_net}
\end{table}
\begin{table}[ht]
\newcommand{\first}{\cellcolor{red!40}}
\centering
\begin{tabular}{lccccc}
\toprule

Model & ATE & RTE & Drift & AYE \\
&($m$)&($m$)&&($deg$)\\
\midrule
TLIO & 10.3005 & 3.6263 & 2.9501 & 3.3684\\
+ rot. aug. &  1.6744 & 0.4944 & 1.5526 & 2.7290\\
+ rot. aug. + more layers & 1.6447 & 0.5466 & 1.2767 & 2.7279\\
+ rot. aug. + Non Eq. Frame& 1.4924 & 0.5119 & 1.2721 & 2.7109\\
+ rot. aug. + PCA Frame & 8.5787 & 2.9962 & 2.0872 & 3.0183\\
\hline
+ \textbf{$\mathbf{SO(2)}$ Eq. Frame } & 1.4850 & 0.4901 & 1.3029 & 2.7615\\
+ \textbf{$\mathbf{O(2)}$ Eq. Frame } & \first{1.4316} & \first{0.4592} & \first{1.3096} & \first{2.7250}  \\
\bottomrule
\end{tabular}
\caption{Results of evaluation of EKF with Random Rotation Transformations (4 per trajectory) on TLIO test dataset (\emph{i.e.}, results on augmentated test dataset). A lower error indicates a better model. The lowest values are annotated with Red. Our proposed methods are in bold.}
\label{aug_ekf}
\end{table}

\subsection{Visualization of RONIN}
\begin{figure}[ht]
\begin{center}
\centerline{\includegraphics[width=\columnwidth]{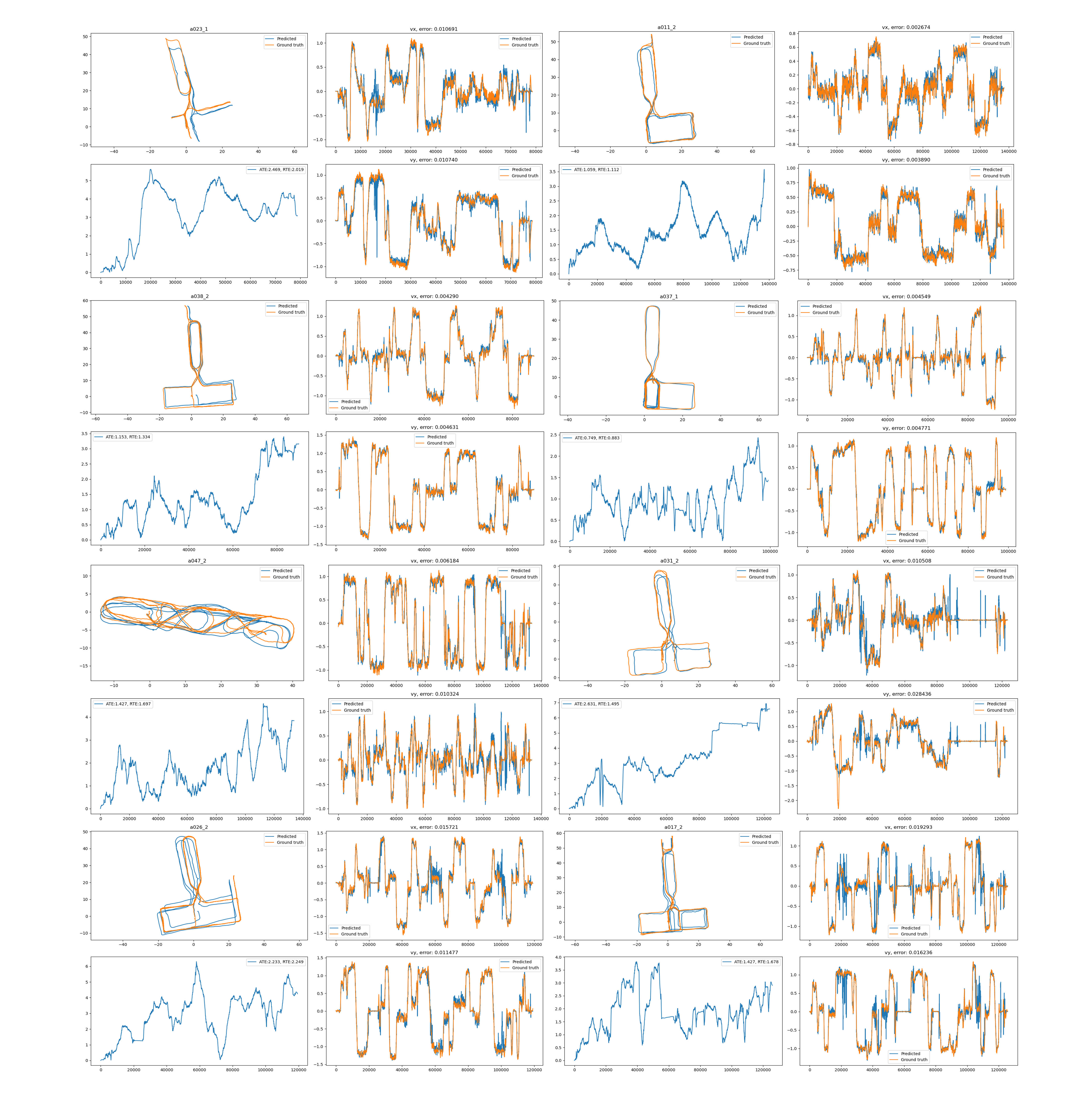}}
\caption{Visualization of RONIN Unseen Test Dataset Trajectories for our best method applied to RONIN, +$O(2)$ Eq. Frame.}
\label{app_ronin}
\end{center}
\end{figure}

The visualization of trajectories in RONIN is displayed in Figure~\ref{app_ronin}.
\subsection{Covariance Consistency}
\label{covariance_consistency}
\begin{figure}[!htb]
\begin{center}
\centerline{\includegraphics[width=0.8\columnwidth]{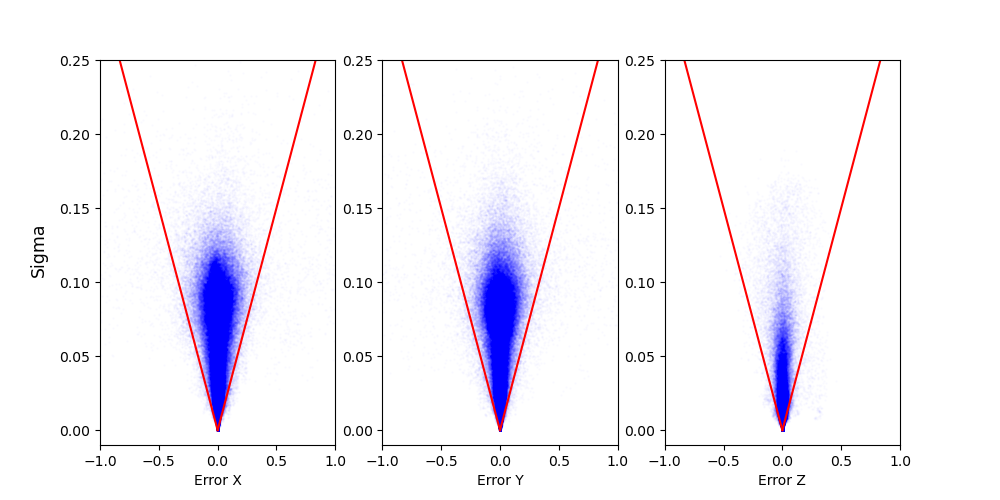}}
\caption{Consistency of Covariance Prediction in the Invariant Space for TLIO test dataset}
\label{cov_fc_tlio}
\end{center}
\end{figure}
\begin{figure}[!htb]
\begin{center}
\centerline{\includegraphics[width=0.8\columnwidth]{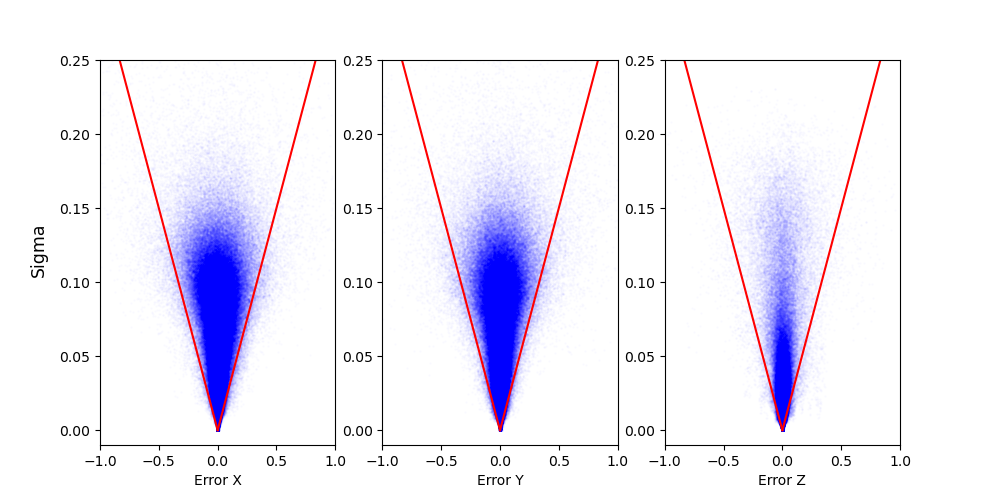}}
\caption{Consistency of Covariance Prediction in the Invariant Space for Aria dataset}
\label{cov_fc}
\end{center}
\end{figure}
Similar to TLIO~\citep{liu2020tlio}, we plot the prediction error against standard deviation ($\sigma$) predicted by the network in the invariant space. As seen in Figure~\ref{cov_fc} and Figure~\ref{cov_fc_tlio} the covariance prediction of our method is consistently within the 3-$\sigma$ depicted by the red lines. These results show that our diagonal covariance prediction in the invariant space is consistent. 

\subsection{Ablation on IMU sequence length}
\label{sequence_length_ablation}
We aligned the sequence length with baseline models for fair comparison. However, in this Section, we ablate on the sequence length as shown in Table~\ref{displacement_window_ablation} and Table~\ref{context_window_ablation}. Table~\ref{displacement_window_ablation} varies sequence lengths and displacement prediction windows (e.g., 0.5s displacement with 0.5s of 200Hz IMU data results in a sequence length of 100). Table~\ref{context_window_ablation} fixes the prediction window at 1s and varies the context window (e.g., a 2s context window with 200Hz IMU data results in a sequence length of 400). Our results confirm TLIO~\citep{liu2020tlio} Sec. VII A.1: increasing the context window reduces MSE but not ATE. A lower MSE loss over the same displacement window does not translate to a lower ATE. Thus, the addition of the equivariant framework does not change the characteristics of the base (off-the-shelf) model used.
\begin{table}[ht]
\newcommand{\first}{\cellcolor{red!40}}
\newcommand{\second}{\cellcolor{orange!40}}
\newcommand{\third}{\cellcolor{yellow!40}}
\centering
\begin{small}
\begin{tabular}{llcccccc}
\toprule
& &\multicolumn{3}{c}{TLIO Dataset} & \multicolumn{3}{c}{Aria Dataset}\\
\cmidrule(lr){3-5} \cmidrule(lr){6-8} 
Model & Context  & MSE* &ATE* &RTE* & MSE* &ATE* &RTE* \\
TLIO&Window (s)&($10^{-2}m^2$)&($m$)&($m$)&($10^{-2}m^2$)&($m$)&($m$)\\
\midrule
+ rot. aug.         &0.5 & 1.132 & 2.029 & 0.340 & 1.038 & 1.489 & 0.332\\
+ rot. aug.        &1 & 3.242 & 3.722 & 0.551 & 5.322 & 2.103 & 0.521\\
+ rot. aug.        &2& 9.862 & 5.102 & 0.944 & 6.717 & 3.452 & 0.970\\

+ $SO(2)$ Eq. Frame   &0.5 & 1.124 & 0.711 & 0.175 & 1.040 & 0.673 & 0.190\\
+ \textbf{$\mathbf{SO(2)}$ Eq. Frame }  &1 & 3.194 & 2.401 & 0.501 & 2.457 & 1.864 & 0.484\\
+ $SO(2)$ Eq. Frame   &2 & 10.019 & 3.862 & 0.797 & 6.569 & 2.745 & 0.774\\

+ $O(2)$ Eq. Frame  &0.5 & 1.040 & 0.595 & 0.136 & 1.002 & 0.589 & 0.148\\
+ \textbf{$\mathbf{O(2)}$ Eq. Frame } &1 & 2.982 & 2.382 & 0.479 & 2.304 & 1.849 & 0.465\\
+ $O(2)$ Eq. Frame  &2 & 9.804 & 4.268 & 0.762 & 6.112 & 2.556 & 0.709\\
\bottomrule
\end{tabular}
\caption{Results for ablation on changing prediction displacement window on TLIO architecture.}
\label{displacement_window_ablation}
\end{small}

\end{table}
\begin{table}[ht]
\newcommand{\first}{\cellcolor{red!40}}
\newcommand{\second}{\cellcolor{orange!40}}
\newcommand{\third}{\cellcolor{yellow!40}}
\centering
\begin{small}
\begin{tabular}{llcccccc}
\toprule
& &\multicolumn{3}{c}{TLIO Dataset} & \multicolumn{3}{c}{Aria Dataset}\\
\cmidrule(lr){3-5} \cmidrule(lr){6-8}
Model & Context & MSE* &ATE* &RTE* & MSE* &ATE* &RTE* \\
TLIO&Window ($s$)&($10^{-2}m^2$)&($m$)&($m$)&($10^{-2}m^2$)&($m$)&($m$)\\
\midrule
+ rot. aug.         &1 & 3.242 & 3.722 & 0.551 & 5.322 & 2.103 & 0.521\\
+ rot. aug.          &2& 3.199 & 2.555 & 0.511 & 3.790 & 2.895 & 0.713\\
+ rot. aug.          &3& 3.284 & 4.463 & 0.617 & 3.511 & 3.014 & 0.738\\

+ \textbf{$\mathbf{SO(2)}$ Eq. Frame }  &1& 3.194 & 2.401 & 0.501 & 2.457 & 1.864 & 0.484\\
+ $SO(2)$ Eq. Frame    &2& 2.886 & 1.837 & 0.429 & 2.187 & 1.533 & 0.444\\
+$SO(2)$ Eq. Frame    &3& 2.790 & 3.090 & 0.492 & 1.986 & 1.684 & 0.447\\

+ \textbf{$\mathbf{O(2)}$ Eq. Frame }&1 & 2.982 & 2.382 & 0.479 & 2.304 & 1.849 & 0.465\\
+ $O(2)$ Eq. Frame  &2 & 2.382 & 1.895 & 0.367 & 1.307 & 1.382 & 0.338 \\
+ $O(2)$ Eq. Frame &3 & 2.161 & 2.083 & 0.366 & 0.974 & 1.672 & 0.366 \\
\bottomrule
\end{tabular}

\caption{Results for ablation on changing context window with fixed displacement window of 1s on TLIO architecture.}
\label{context_window_ablation}
\end{small}

\end{table}

\subsection{Sensitivity analysis to gravity direction perturbation}
\label{gravity_direction_ablation_section}
\begin{table}[ht]
\newcommand{\first}{\cellcolor{red!40}}
\newcommand{\second}{\cellcolor{orange!40}}
\newcommand{\third}{\cellcolor{yellow!40}}
\centering
\begin{small}
\begin{tabular}{p{2.8cm}p{3cm}p{.94cm}p{.54cm}p{.54cm}}
\toprule
 &\multicolumn{3}{c}{TLIO Dataset}  & \\
\cmidrule(lr){3-5} \
Model & $\vec{g}$\, direction perturbation & MSE* &ATE* &RTE* \\
(TLIO)&($deg$)&($10^{-2}m^2$)&($m$)&($m$)\\
\midrule

+ \textbf{$\mathbf{SO(2)}$ Eq. Frame }  &0 & 3.194 & 2.401 & 0.501\\
+ $SO(2)$ Eq. Frame    &2 & 3.201 & 2.409 & 0.500 \\
+ $SO(2)$ Eq. Frame   &4 & 3.206 & 2.404 & 0.498 \\
+ $SO(2)$ Eq. Frame    &6 & 3.241 & 2.442 & 0.501\\
+ $SO(2)$ Eq. Frame    &8 & 3.298 & 2.502 & 0.506\\
+ $O(2)$ Eq. Frame $\ddagger$  &0 & 2.982 & 2.406 & 0.478\\
+ $O(2)$ Eq. Frame $\ddagger$  &2 & 3.198 & 2.663 & 0.498\\
+ $O(2)$ Eq. Frame $\ddagger$  &4 & 3.742 & 3.292 & 0.559\\
+ $O(2)$ Eq. Frame $\ddagger$   &6 & 4.505 & 4.228 & 0.659\\
+ $O(2)$ Eq. Frame $\ddagger$   &8 & 5.433 & 5.218 & 0.768\\
+ \textbf{$\mathbf{O(2)}$ Eq. Frame }  &0 & 2.981 & 2.382 & 0.479\\
+ $O(2)$ Eq. Frame   &2 & 2.988 & 2.390 & 0.480\\
+ $O(2)$ Eq. Frame   &4 & 3.010 & 2.415 & 0.486\\
+ $O(2)$ Eq. Frame   &6 & 3.060 & 2.471 & 0.486\\
+ $O(2)$ Eq. Frame   &8 & 3.095 & 2.506 & 0.489\\
\bottomrule
\end{tabular}
\caption{Results for ablation on changing prediction displacement window on TLIO architecture. \textbf{$\ddagger$}\ implies the network was trained without gravity direction perturbation.}
\label{gravity_dir_sensitivity}
\end{small}

\end{table}

Similar to ~\citet{wang2023surprisingeffectivenessequivariantmodels}, which indicates that the equivariance of $SO(2)$ can even help the rotation around another axis which is close to $z$, we believe that embedding equivariance wouldn’t harm the performance of the model when there is a slight perturbation which is inline with the experimental results as seen in Table~\ref{gravity_dir_sensitivity}.

Table~\ref{gravity_dir_sensitivity} presents the sensitivity analysis to gravity direction perturbation, applied for 5 different ranges, i.e., for 2\textdegree, the gravity direction perturbation of (-2\textdegree, 2\textdegree) is applied to the test dataset. We also present results for + $O(2)$ Eq. Frame model trained without the gravity direction perturbation of (-5\textdegree,5\textdegree) during training. We observe the same trend of stability in MSE* as reported in TLIO~\citep{liu2020tlio} when trained with gravity direction perturbation.

\end{document}